\documentclass[]{oppo}
\usepackage{float}

\usepackage{times}
\usepackage{latexsym}
\usepackage{listings}
\usepackage[T1]{fontenc}
\usepackage[utf8]{inputenc}

\usepackage{microtype}

\usepackage{inconsolata}

\usepackage{graphicx}  % 支持 \resizebox
\usepackage{booktabs}  % 提供 \toprule 等命令
\usepackage{multirow}  % 支持 \multirow
\usepackage{array}     % 扩展表格功能
\usepackage{siunitx}   % 数字列专业对齐
\usepackage[utf8]{inputenc} % allow utf-8 input
\usepackage[T1]{fontenc}    % use 8-bit T1 fonts
\usepackage{hyperref}       % hyperlinks
\usepackage{url}            % simple URL typesetting
\usepackage{booktabs}       % professional-quality tables
\usepackage{amsfonts}       % blackboard math symbols
\usepackage{nicefrac}       % compact symbols for 1/2, etc.
\usepackage{microtype}      % microtypography
\usepackage{xcolor}         % colors
\usepackage{xspace}
\usepackage{fancyhdr}
\usepackage{ifthen}
\usepackage{wrapfig}

\usepackage{etoolbox}
\makeatletter
\patchcmd{\@maketitle}
  {\@title}
  {\@title\\[-0.25ex]\normalsize\bfseries \@author}
  {}{}
\makeatother
\PassOptionsToPackage{breaklinks, hidelinks}{hyperref} 
\usepackage{times}
\usepackage{latexsym}
\usepackage{minted}
\usepackage{placeins} % 导言区
\usepackage[T1]{fontenc}
\usepackage[utf8]{inputenc}
\usepackage{microtype}
\usepackage{inconsolata}
\usepackage{graphicx}
\usepackage{wrapfig}
\usepackage{fontawesome5}
\usepackage{multirow}
\usepackage{booktabs}
\usepackage{makecell}
\usepackage{enumitem}
\usepackage[table,xcdraw]{xcolor}
\usepackage[breaklinks, hidelinks]{hyperref}
\usepackage{colortbl}
\usepackage{amsmath,amssymb} 
\usepackage{lipsum} 
\usepackage{array}
\usepackage{longtable,tabularx}
\usepackage{tcolorbox}
\tcbuselibrary{skins,breakable}
\usepackage{algorithm}
\usepackage{algpseudocode}
\usepackage{amsmath} 
\usepackage{listings}
\usepackage{xcolor}
\usepackage{placeins}
\usepackage{float}
\usepackage{multicol}
\usepackage{siunitx}   % 用于小数点对齐
\usepackage{hyperref}
\usepackage{xcolor}
\usepackage{fancyvrb}

\newcommand{\taxonomy}[1]{\textsc{DEFT}}
\newcommand{\bench}[1]{\textsc{Finder}}
\newcommand{\codebook}{\mathcal{C}}

\usepackage{graphicx}  
\usepackage{tikz}
\usepackage{pgfplots}
\pgfplotsset{compat=1.18}

\setlist[itemize]{topsep=0pt, partopsep=0pt, parsep=0pt, itemsep=1pt, leftmargin=*}
\title{How Far Are We from Genuinely Useful Deep Research Agents?}
\author{%
  \textbf{OPPO AI Agent Team} %
}
\vspace{-6ex}
\abstract{
Deep Research Agents (DRAs) aim to automatically produce analyst-level reports through iterative information retrieval and synthesis. However, most existing DRAs were validated on question-answering benchmarks, while research on generating comprehensive report remains overlooked. Worse, current benchmarks for report synthesis suffers from task complexity and subjective metrics—this fails to reflect user demands and limits the practical utility of generated reports. To address these gaps, we present Fine-grained DEepResearch bench (\bench{}), an enhanced benchmark consisting of 100 human-curated research tasks with 419 structured checklist items that standardize report structure, analytical depth, and factual grounding. Based on approximately 1,000 reports produced by mainstream DRAs, we further propose Deep rEsearch Failure Taxonomy (\taxonomy{}), the first failure taxonomy for deep research agents. DEFT contains 14 fine-grained failure modes across reasoning, retrieval, and generation, and is built upon grounded theory with human–LLM co-annotating and inter-annotator reliability validation. Our experimental findings reveal that current DRAs struggle not with task comprehension but with evidence integration, verification, and reasoning-resilient planning.

\vspace{10pt}
\textbf{Date}: \today

\textbf{Code \& Data \faGithub}: 
\url{https://github.com/OPPO-PersonalAI/FINDER_DEFT}

\textbf{Correspondence}: Wangchunshu Zhou at \url{zhouwangchunshu@oppo.com}, Jiaheng Liu at \url{liujiaheng@nju.edu.cn}
}

\begin{document}
\maketitle
\section{Introduction}

% \begin{wrapfigure}{r}{0.478\textwidth} % {placement}{width}
%     % 'r' = 靠右, 'l' = 靠左
%     % 你可能需要调整顶部的垂直间距，可以尝试取消下面这行的注释：
%     \vspace{-15pt} 
%     \centering
%     % 将图片宽度设置为 \linewidth，使其自动适应 wrapfigure 定义的宽度
%     \includegraphics[width=\linewidth]{figure/drb_finder.pdf}
%     \caption{Comparison between DeepResearch Bench (DRB) and our \bench{}.}
%     \label{fig:example}
%     % \vspace{-3mm}
% \end{wrapfigure}

Deep Research Agents (DRAs) have recently attracted increasing attention due to their ability to autonomously retrieve, analyze, and synthesize web-scale information into structured research reports~\citep{google-gemini-deep-research, openai-deep-research, perplexity-deep-research}. These agents utilize advanced techniques in multi-step web exploration, data retrieval, and synthesis to produce comprehensive reports that would traditionally require hours of manual effort. DRAs are increasingly applied in commercial sectors such as academic research, business intelligence, and knowledge management~\citep{huang2025deep, xu2025comprehensive}. 
% \begin{figure}[htbp]
%     \centering
%     \includegraphics[width=0.478\textwidth]{figure/drb_finder.pdf}
%     \caption{Comparison between DeepResearch Bench (DRB) and our \bench{}.}
%     \label{fig:example}
%     \vspace{-3mm}    
% \end{figure}

However, despite their promising application potential, DRAs still fall short of expectations in real-world report generation tasks~\citep{gou2025mind2web, coelho2025deepresearchgym, patel2025deepscholar, abaskohi2025drbench,liang2025personalizeddeepresearchbenchmarks}. Existing benchmarks are mostly tailored for question-answering (QA)~\citep{wu2025webwalker, wei2025browsecomp, bosse2025deep, chen2025browsecomp-plus} or other types of close-ended tasks~\citep{java2025characterizing}, fail to fully capture the nuances and strict requirements of practical deep research scenarios—where higher standards are imposed on the quality, accuracy, depth, and logical coherence of generated reports. Although a considerable number of open-ended benchmarks currently exist~\citep{du2025deepresearchbenchcomprehensivebenchmark,gou2025mind2web, coelho2025deepresearchgym, patel2025deepscholar, abaskohi2025drbench}, their tasks often stem from LLM-driven sampling or synthesis, leading to deviations from human demands and insufficient complexity. 
\clearpage
\begin{wrapfigure}{r}{0.478\textwidth} % {placement}{width}
    % 'r' = 靠右, 'l' = 靠左
    % 你可能需要调整顶部的垂直间距，可以尝试取消下面这行的注释：
    % \vspace{-15pt} 
    \centering
    % 将图片宽度设置为 \linewidth，使其自动适应 wrapfigure 定义的宽度
    \includegraphics[width=\linewidth]{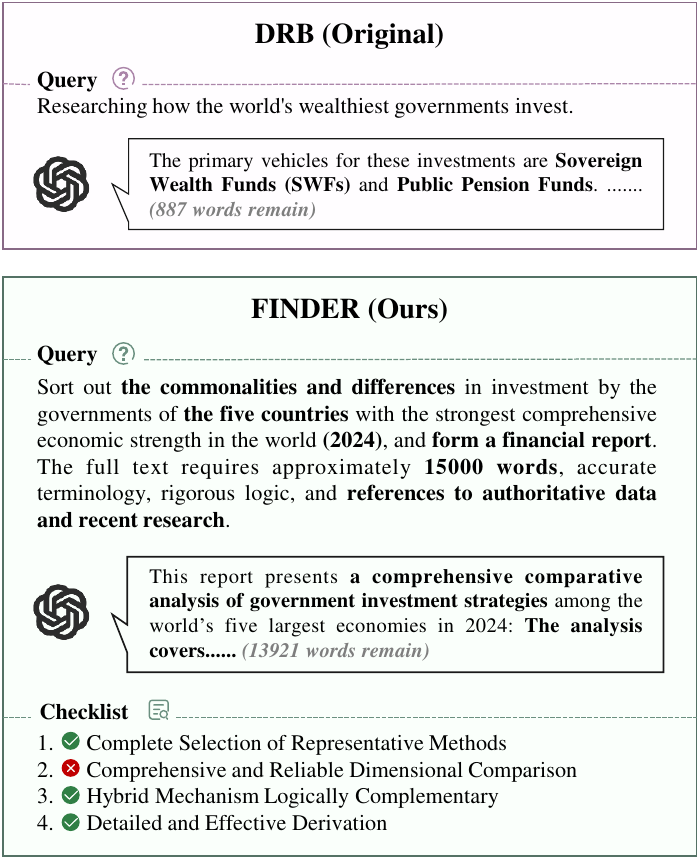}
    \caption{Comparison between DeepResearch Bench (DRB) and our \bench{}.}
    \label{fig:example}
    % \vspace{-3mm}
\end{wrapfigure}
To address this gap, we introduce Fine-grained DEepResearch bench (\bench{}), a fine-grained benchmark designed to evaluate DRAs in a more comprehensive manner. Unlike existing benchmarks, \taxonomy{} is built upon 100 expert-curated research tasks with 419 detailed checklist items that guide the structure, analytical depth, and citation integrity of generated reports. As depicted in \autoref{fig:example}, this explicit guidance enables more structured and reproducible evaluations of the task performance of DRAs. In addition, we propose Deep rEsearch Failure Taxonomy (\taxonomy{}), the first failure taxonomy developed specifically for DRAs. DEFT categorizes common errors into 14 fine-grained failure modes across three core dimensions—reasoning, retrieval, and generation—which we derive through grounded theory~\citep{makri2021grounded,glaser1967discovery} from extensive analysis of ~1,000 generated reports. This taxonomy provides a robust framework for diagnosing where DRAs fail in their reasoning, information seeking, and content generation processes.

Our experimental evaluation on \bench{} and \taxonomy{} of various DRAs, including proprietary systems~\citep{google-gemini-deep-research, openai-deep-research,perplexity-deep-research}, open-source models~\citep{li2025webthinkerempoweringlargereasoning, li2025chainofagentsendtoendagentfoundation, 2025mirothinker,chen2025a2fmadaptiveagentfoundation, qin2025flashsearcherfasteffectiveweb,shi2025taskcraftautomatedgenerationagentic}, and agent frameworks~\citep{2025mirothinker, hu2025owloptimizedworkforcelearning, openmanus2025,zhou2023agents,zhou2024agents2,zhu2025oagentsempiricalstudybuilding,wang2025efficientagentsbuildingeffective,zhu2025scalingtesttimecomputellm,tang2025agent}, reveals several key insights. While systems like Gemini perform well across general benchmarks, our analysis shows that over 39\% of failures arise in content generation, particularly through strategic content fabrication, where agents tend to generate unsupported but seemingly professional content. Furthermore, retrieval-related failures, such as insufficient evidence integration and fact-checking issues, account for over 32\% of errors, highlighting the challenges DRAs face in managing and verifying the quality of retrieved information. These results underscore that the core challenges for DRAs are not limited to simple task comprehension but instead involve deeper issues in evidence verification and reasoning resilience. To summarize, our contributions are as follows,

\begin{itemize}  
     \item We propose \bench{}, a fine-grained benchmark with 100 expert-curated tasks and 419 structured checklist items, enabling robust and reproducible evaluation of DRAs across various dimensions of research report generation.
    \item We establish \taxonomy{}, the first failure taxonomy for DRAs, which categorizes errors into 14 fine-grained failure modes under three core dimensions: reasoning, retrieval, and generation.
    \item Through experiments on proprietary APIs, open-source models, and agent frameworks, we demonstrate that current DRAs struggle more with evidence integration and methodological rigor than with understanding tasks, revealing key weaknesses in reasoning resilience and strategic content fabrication.
\end{itemize}

\section{Related Works}
\enlargethispage{\baselineskip}
% \vspace{-0.01mm}
% \subsection{Deep Research Agents}
% Deep research agents are designed to execute multi-turn knowledge-intensive tasks utilizing global planning and iterative tool use. While the concept was academically formalized in 2025, commercial deployment was rapidly pioneered by Google and OpenAI, positioning DRAs as the successor to standard Retrieval Augmented Generation (RAG). Existing work achieves deep research through agentic training, agentic engineering, or a combination of both.
% \subsection{Deep Research Benchmarks}

Early works on DRAs~\citep{google-gemini-deep-research, openai-deep-research,perplexity-deep-research} employed datasets towards AGI as evaluation benchmarks. The most representative examples include GAIA~\citep{mialon2023gaia} and HLE~\citep{phan2025hle}. As the deep research community grows, researchers have proposed various specialized benchmarks~\citep{wu2025webwalker, wei2025browsecomp, bosse2025deep, java2025characterizing,chen2025browsecomp-plus}. Although the aforementioned datasets are challenging, they all fall under \textbf{closed-ended} assessments with standard answers. They neglect the evaluation of report generation, exhibiting a mismatch with the requirements of deep research. In contrast, open-ended benchmarks treat deep research as a task without a single definitive solution. DeepResearch Bench \citep{du2025deepresearchbenchcomprehensivebenchmark} contains 100 PhD-level problems spanning 22 domains, introducing the RACE and FACT evaluations for report quality and effectiveness. Mind2Web 2 \citep{gou2025mind2web} comprises 130 time-varying daily life tasks and proposes an ``Agent-as-Judge'' framework to achieve automated  verification and attribution. DeepResearchGym \citep{coelho2025deepresearchgym} provides sandbox environments with reproducible search APIs and standardized protocols for transparent deep research benchmarking. DeepScholar-Bench ~\citep{patel2025deepscholar} is a benchmark that automatically evaluates  research synthesis abilities through content coverage, citation accuracy, and organizational quality. DRBench~\citep{abaskohi2025drbench} focuses on enterprise scenarios and evaluates DRAs through judge-based, citation-grounded assessment of long-form analytical reports. However, due to the dynamic nature of research reports, all these benchmarks employ subjective metrics based on the authors' experience or domain knowledge. Different benchmarks utilize varying metrics, lacking a unified standard.
\raggedbottom
\section{Methodology}
% \begin{figure*}[!tb]
%     \centering
%     \includegraphics[width=1\textwidth]{figure/DRAST_PIPLINE_V2.pdf}
%     \caption{Overview of the \taxonomy{} Failure Taxonomy Construction Framework.}
%     \label{fig:DRAST_pipline}
%     \vspace{-3mm}
% \end{figure*}

% \begin{figure*}[!tb]
%     \centering
%     \includegraphics[width=1\textwidth]{figure/taxonomy_build_v2.pdf}
%     \caption{Overview of the \taxonomy{} Construction.}
%     \label{fig:taxonomy_build}
%     \vspace{-3mm}
% \end{figure*}
\begin{figure*}[!htbp]
    \centering
    \includegraphics[width=1\textwidth]{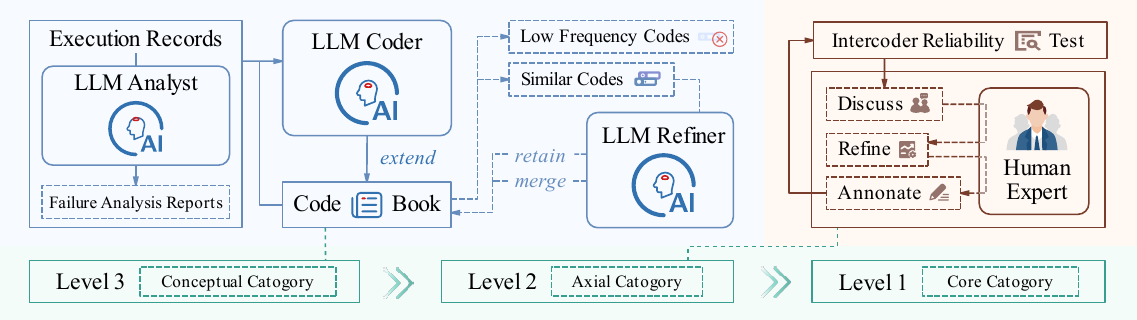}
    \caption{Overview of the \taxonomy{} Construction.}
    \label{fig:taxonomy_build}
    \vspace{-3mm}
\end{figure*}

% In this section, we present the comprehensive methodology employed to construct \taxonomy{}, as illustrated in \autoref{fig:taxonomy_build}. To capture and analyze agent failures in complex research tasks, we first establish \bench{}, a fine-grained version of DeepResearch Bench\cite{du2025deepresearchbenchcomprehensivebenchmark}(\S~\ref{sec:bench}). Leveraging experimental data collected from the \bench{}, we employ grounded theory to construct through a three-stage human-machine collaborative coding process(\S~\ref{sec:taxonomy}). 

\subsection{Fine-grained DEepResearch bench (\bench{})} \label{sec:bench}
Based on DeepResearch Bench, we refine the prompts and add structured checklists to construct \bench{}, aiming to enhance evaluation precision and reproducibility.

\subsubsection{Preliminary : DeepResearch Bench}

The DeepResearch Bench consists of 100 PhD-level research tasks (50 in Chinese and 50 in English) designed to evaluate Deep Research Agents (DRA). It introduces two evaluation frameworks: RACE, which dynamically scores report quality in terms of comprehensiveness, depth, instruction-following, and readability; and FACT, which assesses retrieval effectiveness through citation accuracy and average effective citations (see \autoref{app:deepresearch_bench_framework} for detailed description). While DeepResearch Bench offers robust metrics for report evaluation, focusing solely on the final report does not adequately reflect a model’s reasoning seach and information seeking capabilities that underpin its deep research performance.

% While DeepResearch Bench provides robust metrics for evaluating report-generation quality, report quality alone does not fully capture a model’s capability for real deep research.

% Report quality reflects the final output, whereas deep research capability should also encompass a model’s performance throughout the report generation process. Furthermore, the dataset of DeepResearch Bench is filtered from dialogues between large language models and real users. Most of these prompts are brief, typically consisting of one or two sentences, and thus fail to meet the requirements for generating high-quality reports.

\subsubsection{Prompt Refinement}
To address the issue of overly brief queries in the DeepResearch Bench, we invited seven domain experts to expand the queries in the DeepResearch Bench according to their respective areas of expertise. For each query, explicit guidelines were established regarding the report’s length, disciplinary scope, presentation format, and other aspects. To ensure the correct generation type, each report was required to include the term ``report'' or equivalent expressions. Additionally, an independent expert who was not involved in the rewriting process manually evaluated the quality of the revised outputs. The finalized queries are presented in \autoref{fig:example}. As shown in \autoref{fig:word count}, our queries are longer than those in the original DeepResearch Bench. While preserving the independent semantic integrity of each sentence, the increased query length signifies a higher degree of task specification and research complexity.

\subsubsection{Checklist Construction} \label{sec:checklist design}

% To ensure consistency in evaluation, we adopted a concise three-step process for checklist construction.

% \paragraph{Expert drafting.}
% The human experts created five checklists for each prompt.

% \paragraph{LLM-based relevance filtering.} \label{step:llm_filter}
% We used Gemini 2.5 Flash~\cite{2025arXiv250706261C} to compute the semantic relevance between each checklist item and its prompt. 
% The relevance score $s \in [0,1]$ denotes the model-estimated semantic similarity, and items with $s > 0.8$ were retained.

% \paragraph{Expert cross-validation and iterative refinement.}
% Experts reviewed checklists written by others and removed low-quality or redundant items. The refined checklists were then re-evaluated by the judge model from Step 2. This review–evaluation cycle was repeated until all checklists achieved stable scores above 0.8, ensuring convergence and consistency across evaluations.

% In total, \textbf{419} validated checklists were obtained, with each question associated with about \textbf{3--5 }checklists. The distribution is shown in \autoref{fig:checklist_distribution}.

To make the evaluation more structured, experts were first required to create five checklists for each query based on its specific characteristics. Each checklist served two purposes: first, to organize and structure the existing information within the query, and second, to supplement additional content requirements and constraints that were not explicitly mentioned but were relevant to the query. This approach ensured that the checklists were comprehensive and systematic during the evaluation process.

Subsequently, we used the Gemini 2.5 Flash to refine the initially generated checklists by eliminating items with incomplete semantics, ambiguous expressions, or those irrelevant to the reports generated for the corresponding queries. This process was conducted iteratively until all checklists met the predefined standards.

In total, we generated \textbf{419} checklists for 100 queries, with each query containing between three and five checklists. The distribution of checklist numbers is presented in \autoref{fig:checklist_distribution}. Further examples of queries are provided in Appendix~\ref{app:query_examples}.

\subsection{Failure Taxonomy} \label{sec:taxonomy}

We construct a comprehensive failure taxonomy to systematically identify, categorize, and interpret the underlying causes of Deep Research Agent (DRA) errors. To avoid the subjective biases and omissions that may arise when relying solely on researchers’ intuition or prior literature, the taxonomy is developed through a human-AI collaborative framework comprising open (\S~\ref{sec:open coding}), axial (\S~\ref{sec:axial coding}), and selective coding (\S~\ref{sec:selective coding}). The design of this process draws on grounded theory, which is a classic qualitative methodology that has been widely adopted across disciplines such as management \cite{makri2021grounded}, education \cite{stough2021grounded}, and software engineering \cite{hoda2021socio} to construct evaluation or attribution schemata. The entire procedure (\autoref{fig:taxonomy_build}) has been formalized into a pseudocode workflow, which is presented in \autoref{appendix:taxonomy-pipeline}.

% Another version: The design of this process draws on grounded theory, which is a classic qualitative methodology that inductively generates theory from empirical data rather than testing preconceived hypotheses.

\subsubsection{Open Coding} \label{sec:open coding}

\paragraph{Conceptual Category Generation.}
% \textbf{Conceptual Category Generation.}

Open coding entails analyzing and conceptualizing raw textual data to identify and label underlying conceptual categories within the study context \cite{gridach2025agenticaiscientificdiscovery}. Specifically, we collected performance metrics for \textbf{nine} deep research agent tasks in our benchmark and selected \textbf{five} large language models (Claude Opus~4.1, Gemini~2.5~Pro, Grok~4, DeepSeek-V3.1, and Qwen3-Max-Preview) from distinct model families to serve as coders. This design leverages their diverse inductive biases to broaden coverage and enhance coding breadth.

% (Claude Opus~4.1, Gemini~2.5~Pro, Grok~4, DeepSeek-V3.1, and Qwen3-Max-Preview)

To systematically manage the coding process, we adopted the core principle of constant comparative analysis from grounded theory and maintained a dynamically updated conceptual inventory, hereafter referred to as the \textit{codebook}~($\codebook$), defined as:
\begin{equation}
\codebook = \bigl\{\, (c_i, d_i) \;\big|\; i = 1,2,\ldots,N \,\bigr\},
\label{eq:codebook-def}
\end{equation}
where $c_i$ denotes the concept name and $d_i$ its corresponding brief textual description. For each new concept identified, we first attempt to match it with existing $c_i$; if no match is found, a new pair $(c_{N+1}, d_{N+1})$ is added to $\codebook$.

Additionally, to focus the model on identifying and labeling failure modes rather than conducting deep causal analysis or automated failure localization\citep{zhang2025agentracer}, we instructed it to first generate a failure analysis report (\autoref{app: Failure Report} shows an example of the report) for each execution case as supplementary material to the original coding data. During the initial coding phase, we established a set of seed concepts (\autoref{app:seed}) based on the research findings of \citet{tang2025agent} and \citet{cemri2025multi} to construct few-shot prompts that guided the large language model's coding process.

\paragraph{Conceptual Category Optimization.}
% \textbf{Conceptual Category Optimization.}
Whether within the same LLM coder or across multiple LLM coders, generated codes may exhibit redundancy or outliers. We address this through category optimization. On one hand, we employ Seed1.5-Embedding, which ranked first in MTEB  (eng-v2, API available) \citep{enevoldsen2025mmteb}, as the embedding model to identify concept pairs with cosine similarity $\geq 0.6$. These pairs are then fed into the large language model to be merged where appropriate. Additionally, concepts appearing below a removal threshold are discarded. As shown in \autoref{tab:group_comparison}, we divided the source material into two groups for open-ended coding, each undergoing two rounds of refinement. The first round was conducted independently by each LLM coder, while the second round integrated the coding results from five LLM coders. An additional refinement round consolidated the coding outcomes between the two groups, ultimately yielding 51 concepts.

\subsubsection{Axial Coding} \label{sec:axial coding}

Axial coding employs both deductive and inductive reasoning to explore relationships among concepts based on semantics, context, process, causality, function, structure, and strategy \citep{hoda2024qualitative}. Through merging, splitting, removing, or modifying these relationships, it forms axial categories. At this stage, we conducted three rounds of coding based on inter-coder reliability (ICR) assessments: the first round utilized open coding results from Group~A (\autoref{tab:group_comparison}), while the second and third rounds incorporated all open coding results alongside the first-round axial coding outcomes. ICR measures the consistency among coders when encoding the same data \citep{o2020intercoder} and has been demonstrated to consolidate \citep{olson2016applying, diaz2023applying} or validate \citep{nili2020approach} existing coding frameworks. We selected Krippendorff's Alpha \citep{krippendorff2018content} to assess ICR. The universal formula for Krippendorff's Alpha is as follows:
\begin{equation}
\alpha = 1 - \frac{D_o}{D_e}
\end{equation}
where $D_o$ denotes observed disagreement and $D_e$ denotes expected disagreement by chance. For practical calculations, we utilized the web-based statistical package K-Alpha Calculator \citep{marzi2024k}.

Following each coding round, to conduct ICR assessment, we engaged three domain experts to independently annotate a randomly sampled subset. This subset comprised 24 (first round) or 54 execution records (second and third rounds), with 3 logs selected from each Chinese and English version of each framework. It takes approximately 5~hours for experts to engage in discussion following each annotation round to resolve discrepancies and refine category definitions. After a few iterations, we finalized 14 axial categories. Detailed definitions of each category are provided in \autoref{Axial Category Definitions}, and illustrative case studies for each category are presented in \autoref{app: Taxonomy Case Study}.

\subsubsection{Selective Coding} \label{sec:selective coding}

Selective coding synthesizes the concepts and categories developed in the first two coding stages to establish overarching core categories. It clarifies their interrelationships and connects them through systematic logical threads \citep{makri2021grounded}. At this stage, we repeatedly analyzed the axial categories derived from axial coding, ultimately distilling three core categories: \textit{Reasoning}, \textit{Retrieval}, and \textit{Generation}. Functionally, these three core categories form a complete closed-loop for agent task execution. Temporally, they are interwoven and sequentially progressive, collectively underpinning a systematic understanding of agent failure mechanisms.

We randomly selected 36 execution records (six each from the Chinese and English part) generated by two agents not involved in the taxonomy construction stage, WebThinker and OpenManus, for coding analysis. No new categories emerged during this process, indicating that our categorization system had achieved theoretical saturation and demonstrated the explanatory power and stability required by grounded theory \citep{wutich2024sample}.

\subsubsection{Positive Taxonomy Metric}

To establish a unified and success-oriented framework for performance evaluation within the failure-mode taxonomy, we introduce a \textit{positive performance metric} that transforms model error counts in each category into a bounded, interpretable score.

Let $E_i$ denote the number of observed errors in category $i \in \{1, \dots, |\mathcal{T}|\}$, and let $|D|$ represent the total size of the dataset. Inspired by the concept of \textit{cosine similarity} in vector space models~\cite{10.1145/361219.361220}, we define the performance score as
\begin{equation}
S_i = |D| \cdot \cos\!\left(\frac{E_i}{|D|}\cdot\frac{\pi}{2}\right).
\end{equation}

Here, $S_i$ captures the angular deviation of model performance from an error-free optimum. When $E_i = 0$, the model attains the maximum possible score $S_i = |D|$. As the number of errors $E_i$ increases, $S_i$ monotonically decreases toward 0, reflecting a gradual decline in performance. Further justification of this formulation is provided in \autoref{app:cosine-metric}.

\section{Experiments}

% \paragraph{Evaluated Models.}
\subsection{Evaluated Models.}
We evaluate three representative categories of systems.
(1) \emph{\textbf{Proprietary API}} comprise Gemini-2.5-Pro Deep Research, O3 Deep Research, O4-Mini Deep Research, and Perplexity Deep Research, which are closed-source research agents accessible through API interfaces.
(2) \emph{\textbf{Open-source Model}} include open-source or self-hosted reasoning models such as MiroThinker, WebThinker, and AFM.
(3) \emph{\textbf{Agent Framework}} encompass OWL, OpenManus, and MiroFlow, where MiroFlow is evaluated in both English and Chinese versions to examine cross-lingual performance within a unified framework.
Comprehensive model configurations and parameter settings are detailed in \autoref{app:model_config}.

\subsection{\bench{} Performance Analysis}
\label{sec : finder performance}
We evaluate model performance across three dimensions: RACE and FACT, Positive Taxonomy Metrics, and the Checklist Accuracy. The overall outcomes are summarized in \autoref{tab:overall-results}. 
% This section provides a comparative analysis from these three perspectives to reveal strengths and limitations across model categories.

% proprietary APIs, agent frameworks, and open-source models
\begin{table*}[htbp]
\centering
\resizebox{0.99\textwidth}{!}{
\renewcommand{\arraystretch}{1.5}
\begin{tabular}{c c c c c c c c c c c c c}
\hline
\multirow{2}{*}{\textbf{Model}} & \multicolumn{5}{c}{\textbf{RACE}} & \multicolumn{2}{c}{\textbf{FACT}} & \multicolumn{4}{c}{\textbf{Positive Taxonomy Metric}} & \multirow{2}{*}{\makecell{\textbf{Checklist}\\ \textbf{Pass Rate}}} \\
\cmidrule(lr){2-6} \cmidrule(lr){7-8} \cmidrule(lr){9-12}
& Overall & Comp. & Depth & Inst. & Read. & C.Acc. & E.Cit. & Rea. ($S_1$) & Ret. ($S_2$) & Gen. ($S_3$) & $S_{avg}$ & \\
\hline
\multicolumn{13}{c}{\textit{\textbf{Proprietary API}}} \\
\hline

Gemini-2.5-Pro Deep Research\cite{google-gemini-deep-research}  & \textbf{50.95} & \textbf{52.05} & \textbf{49.92} & 50.55 & \textbf{48.51} & 57.09 & 48.38 & 89.80 & \textbf{97.23} & \textbf{89.80} & \textbf{72.89} & 63.01 \\
Kimi K2\cite{kimiteam2025kimik2openagentic}        & 48.28 & 49.60 & 44.77 & \textbf{51.08} & 48.26 & - & - & \textbf{93.54} & 82.71 & 20.28 & 65.51 & \textbf{66.59} \\

O3 Deep Research\cite{openai_o3_deep_research}        & 46.25 & 47.82 & 42.13 & 49.87 & 46.61 & \textbf{65.98} & \textbf{76.58} & 73.96 & 39.71 & 43.99 & 52.56 & 57.52 \\
O4-Mini Deep Research\cite{openai_o4_mini_deep_research}        & 43.49 & 43.91 & 38.00 & 49.21 & 44.02 & - & - & 93.54 & \textbf{75.01} & 45.40 & 71.32 & 56.09 \\
Perplexity Deep Research\cite{perplexity-deep-research}         & 41.62 & 43.68 & 38.39 & 44.30 & 41.12 & 5.25 & 29.31 & 50.90 & 60.04 & 30.90 & 47.28 & 51.55 \\
\hline
\multicolumn{13}{c}{\textit{\textbf{Open-source Model}}} \\
\hline
WebThinker\cite{li2025webthinkerempoweringlargereasoning}      & \textbf{41.11} & \textbf{41.43} & \textbf{34.51} & \textbf{47.71} & \textbf{43.56} & 11.32 & 1.83 & \textbf{72.70} & 24.87 & 9.41 & 35.73 & 44.87 \\
AFM\cite{li2025chainofagentsendtoendagentfoundation}             & 37.97 & 39.69 & 34.92 & 39.17 & 39.93 & 23.80 & \textbf{83.64} & 41.15 & \textbf{57.50} & 18.74 & \textbf{36.86} & 48.45 \\
MiroThinker\cite{miromind2025mirothinker}    & 33.51 & 32.94 & 26.01 & 39.20 & 40.42 & \textbf{41.60} & 1.13 & 50.90 & 26.39 & 15.64 & 30.98 & 50.84 \\
Tongyi-DeepResearch\cite{tongyidr}    & 30.06 & 31.50 & 24.60 & 35.02 & 32.81 & 18.18 & 2.75 & 30.90 & 30.90 & \textbf{46.79} & 36.20 & \textbf{67.54} \\
\hline
\multicolumn{13}{c}{\textit{\textbf{Agent Framework}}} \\
\hline
MiroFlow-English\cite{2025mirothinker}    & \textbf{42.20} & 42.84 & \textbf{36.49} & \textbf{47.55} & \textbf{44.51} & \textbf{22.73} & 2.00 & 54.90 & \textbf{46.79} & 15.64 & \textbf{39.11} & \textbf{72.19} \\
MiroFlow-Chinese\cite{2025mirothinker}   & 41.28 & \textbf{43.25} & 36.11 & 44.92 & 43.63 & 16.67 & 2.47 & 54.90 & 46.79 & 15.64 & 39.11 & 54.80 \\
OWL\cite{hu2025owloptimizedworkforcelearning}            & 39.22 & 39.57 & 33.81 & 44.41 & 40.13 & - & - & 49.55 & 43.99 & \textbf{29.40} & 40.98 & 53.94 \\
OpenManus\cite{openmanus2025}     & 35.44 & 35.23 & 29.02 & 41.95 & 37.50 & 8.84 & \textbf{4.08} & \textbf{62.52} & 33.87 & 18.74 & 38.38 & 61.34 \\
\hline
\end{tabular}
}
% \caption{Overall evaluation results of \textbf{\bench{}}. Evaluations span three complementary modules: RACE, FACT, and our \taxonomy{} Positive Metric (\textit{reasoning} $S_1$, \textit{retrieval} $S_2$, and \textit{Generation} $S_3$). The final column reports the Checklist Pass Rate. ``–'' indicate unavailable results due to framework limitations in citation extraction or verification. \textbf{Bold} values denote the highest score within each group.}

\caption{Overall evaluation results of \textbf{\bench{}} across three complementary modules: RACE, FACT, and our \taxonomy{} Positive Metric (\textit{reasoning} $S_1$, \textit{retrieval} $S_2$, and \textit{generation} $S_3$). The final column reports the Checklist Pass Rate. ``–'' indicates missing or unavailable results; detailed explanations of these cases are provided in Appendix~\ref{appendix:fact_failure}. \textbf{Bold} values denote the highest score within each group.}

\vspace{-3mm}
\label{tab:overall-results}
\end{table*}

\paragraph{RACE and FACT.}
Under the RACE framework, Gemini 2.5 Pro Deep Research remains the top performer with an overall score of 50.95, followed by Kimi K2 (48.28) and O3 Deep Research (46.25). Among the Open-source Models and Agent Frameworks, WebThinker and MiroFlow stand out for their strong instruction adherence. MiroFlow was further evaluated using English and Chinese prompts from FINDER, each repeated three times to mitigate randomness; the results show that English tasks achieved slightly higher scores (42.20) compared to the Chinese version (41.28), indicating superior reasoning and text organization in English. Within the FACT framework, O3 Deep Research demonstrates exceptional performance, leading significantly in both factual precision (65.98) and citation reliability (76.58), while Gemini 2.5 Pro Deep Research follows as a strong contender, with the lower scores or data gaps for other models likely stemming from the more challenging upgraded prompts that demand denser reasoning and stricter citation validation.

% These two metrics together establish the output quality baseline: RACE captures how well models write, while FACT measures how truthfully they support what they write.

% -----------------------------------------------------------
\paragraph{Positive Taxonomy Metrics.}

% The taxonomy results provide a process-level view of how models reason and synthesize information. Gemini attains balanced high scores across cognition, retrieval, and generation, suggesting well-coordinated task understanding and synthesis. O3 and O4-Mini display strong retrieval but less coherent generation, while open frameworks like MiroFlow maintain moderate but stable profiles, indicating the stabilizing effect of structured orchestration. Overall, these metrics reveal that superior systems maintain equilibrium among understanding, evidence collection, and synthesis—rather than over-optimizing any single phase.

\begin{figure*}[!tb]
    \centering
    \includegraphics[width=1\textwidth]{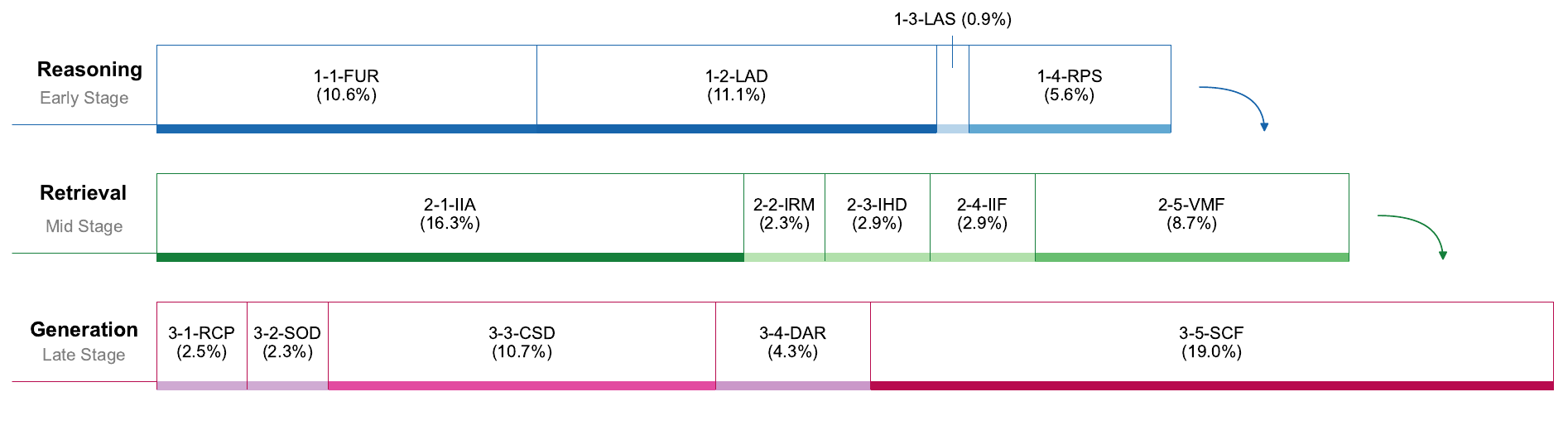}
    \caption{Overview of the Level~1 (Core) and Level~2 (Axial) Failure Categories in \taxonomy{}}
    \label{fig:taxonomy}
    \vspace{-3mm}
\end{figure*}

The taxonomy results offer a process-level perspective on how models reason and synthesize information. Gemini achieves consistently high scores across reasoning , retrieval , and generation , indicating well-coordinated task understanding and synthesis. In contrast, Kimi K2 and O4-Mini exhibit exceptional reasoning capabilities (surpassing Gemini) and strong retrieval performance, but suffer from a sharp decline in generation scores. Open frameworks such as MiroFlow show moderate stability yet similarly face bottlenecks in the final generation stage. Overall, these metrics demonstrate that \textbf{superior systems maintain a balance among understanding, evidence collection, and synthesis rather than overoptimizing a single stage.}

% To further examine system robustness, we analyze the taxonomy-based positive success metrics $S = \{S_1, S_2, S_3\}$ defined in~\S\ref{sec:evaluation}. Cognition-level success ($S_1$) is generally high among API agents, indicating strong semantic understanding and planning capacity for complex tasks. Retrieval-level success ($S_2$) varies more widely, reflecting different strategies for external knowledge utilization. Generation-level robustness ($S_3$) remains the weakest component, revealing a common tendency toward logical drift or hallucination in extended text generation. The averaged score $S_{\text{avg}}$ remains stable in cross-lingual frameworks such as MiroFlow-Chinese, demonstrating that consistent task structures can help preserve reasoning integrity across languages.
% ----------------------------------------------------------

\paragraph{Checklist Accuracy.}

Checklist scores represent meta-reasoning and procedural adherence to the intended research workflow. MiroFlow-English achieves the highest score (72.19), followed by a competitive cluster including Tongyi-DeepResearch (67.54), Kimi K2 (66.59), and Gemini 2.5 Pro (63.01). While MiroFlow demonstrates the specific advantage of explicit agentic orchestration, proprietary models like Kimi and Gemini remain robust, outperforming O3 (57.52) and other baselines. This distribution suggests that \textbf{systematic reasoning discipline—whether through framework design or intrinsic model capability—determines research reliability.}

% The Checklist Pass Rate quantifies explicit instruction-following capability (see~\S\ref{sec:checklist design} and Appendix~\ref{app:checklist}). API-based agents achieve the highest compliance rates, confirming their reliability in structured and template-based tasks. Framework-based systems exhibit moderate scores, reflecting a trade-off between flexibility and adherence. This metric complements RACE and FACT by measuring task-level obedience and offering a more direct view of instruction compliance.

% \paragraph{Overall Observation.}
% Collectively, the three evaluation dimensions outline a multifaceted view of system capability. API agents excel in factual reliability and structural coherence but depend heavily on closed-source retrieval mechanisms. Open frameworks, while transparent and extensible, still face challenges in evidence grounding and precision of task execution. The taxonomy-based analysis further indicates that cognitive proficiency alone does not guarantee factual consistency; robust retrieval grounding remains the key determinant of dependable deep-research performance.

\subsection{DRB vs \bench{}}

\begin{figure*}[htbp]
    \centering
    \includegraphics[width=1.01\textwidth]{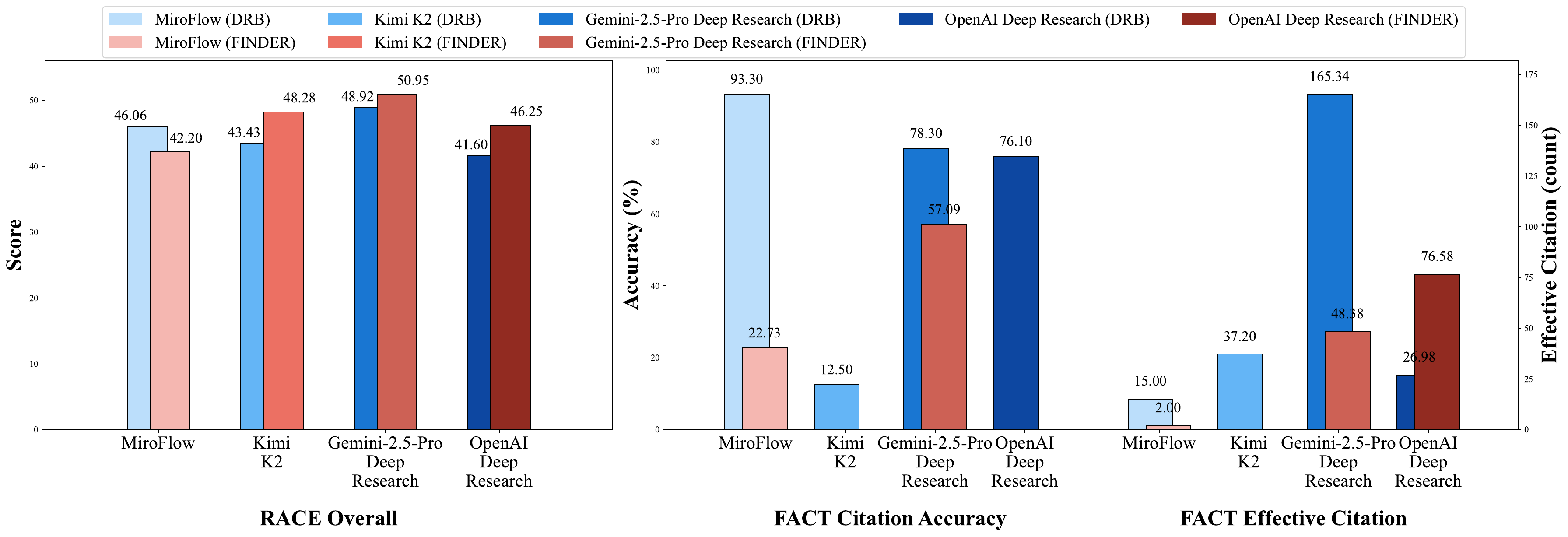}
    \caption{Overview of agent performance on DeepResearch Bench (DRB) and our \bench{}.}
    \label{fig:DRB_vs_FinDer}
\end{figure*}

As shown in \autoref{fig:DRB_vs_FinDer}, we compare the original DeepResearch Bench (DRB) with our proposed \bench{} under both the RACE and FACT frameworks, and this analysis reveals partially divergent outcomes across the two evaluation frameworks.

In the \textbf{RACE} framework, the overall scores under \bench{} remain largely consistent with those from DRB. This consistency arises because both benchmarks share the same reference-based evaluation process: each model’s research report is assessed relative to a standardized reference report (\texttt{reference.jsonl}) generated by Gemini-2.5-Pro Deep Research. The RACE framework evaluates the relative quality of a target report rather than its absolute performance, using four adaptive dimensions(comprehensiveness, depth of insight, instruction-following, and readability). Consequently, differences in absolute RACE scores across benchmarks hold limited interpretive value; only intra-benchmark comparisons among models reliably reflect relative generation quality.

In contrast, the \textbf{FACT} module shows more pronounced disparities between DRB and \bench{}. While OpenAI Deep Research achieves a modest improvement in effective citation (\textit{E.Cit.}), most other systems experience declines in both citation accuracy (\textit{C.Acc.}) and effectiveness. This likely reflects the heightened difficulty introduced by our revised prompt design in \bench{}, which imposes stricter factuality and citation validation demands. The resulting higher citation variance indicates that \bench{} provides a more rigorous test of factual consistency and evidence trustworthiness. Overall, these outcomes suggest that \bench{} enforces stronger constraints on reasoning transparency and source reliability, thereby exposing model weaknesses that were less evident under DRB’s original configuration.

\subsection{Deep Research Failure Taxonomy (\taxonomy{})}

% This section mainly introduces the three Level~1 (core) categories in the \taxonomy{} (\autoref{fig:taxonomy}), and presents some key insights for DRA improvement based on the \taxonomy{} analysis.

This section introduces both the Level 1 (core) and Level 2 (axial) categories in the taxonomy, as shown in \autoref{tab:taxonomy-english-two-cols}. Detailed definitions of axial category are provided in \autoref{Axial Category Definitions}. Furthermore, this section synthesizes key implications for enhancing DRA performance that emerge from the taxonomy-based analysis.

\begin{table}[h!]
\centering
\renewcommand{\arraystretch}{1.2}
\setlength{\tabcolsep}{8pt}
\begin{tabular}{@{}>{\centering\arraybackslash}p{5cm} 
@{\hspace{12pt}} >{\centering\arraybackslash}p{10cm}@{}}
\toprule
\textbf{Level 1 (Core Category)} & \textbf{Level 2 (Axial Category)} \\
\midrule
\multirow{4}{*}{Reasoning}
  & Failure to Understand Requirements (FUR) \\
  & Lack of Analytical Depth (LAD) \\
  & Limited Analytical Scope (LAS) \\
  & Rigid Planning Strategy (RPS) \\
\midrule
\multirow{5}{*}{Retrieval}
  & Insufficient External Information Acquisition (IIA) \\
  & Information Representation Misalignment (IRM) \\
  & Information Handling Deficiency (IHD) \\
  & Information Integration Failure (IIF) \\
  & Verification Mechanism Failure (VMF) \\
\midrule
\multirow{5}{*}{Generation}
  & Redundant Content Piling (RCP) \\
  & Structural Organization Dysfunction (SOD) \\
  & Content Specification Deviation (CSD) \\
  & Deficient Analytical Rigor (DAR) \\
  & Strategic Content Fabrication (SCF) \\
\bottomrule
\end{tabular}
\caption{Taxonomy with Level 1 and Level 2 categories.}
\label{tab:taxonomy-english-two-cols}
\end{table}

\paragraph{Reasoning Category}refers to the failures mainly exhibited during the initial stage of execution, arising from insufficient consideration of user intent or problem details. Specifically, they include Failure to Understand Requirements (1-1-FUR, 10.55\%), Lack of Analytical Depth (1-2-LAD, 11.09\%), Limited Analytical Scope (1-3-LAS, 0.90\%), and Rigid Planning Strategy (1-4-RPS, 5.60\%).

The relatively low proportion of failures in this category indicates that most DRAs are capable of inheriting the underlying large models’ strengths in terms of semantic understanding and basic reasoning \citep{gridach2025agenticaiscientificdiscovery}. However, the issue of 1-4-RPS suggests that the agents still exhibit limitations in dynamic task scheduling and adaptive reasoning. The linear execution logic present in some frameworks often fails to respond effectively to task evolution or intermediate feedback, leading to reduced efficiency or error propagation. In addition, 1-2-LAD and 1-3-LAS represent two orthogonal dimensions of reasoning capability. An ideal deep research agent should possess both strong problem-decomposition skills and robust system-modeling abilities.

\begin{tcolorbox}[
  colback={rgb,255:red,230; green,240; blue,250},
  colframe={rgb,255:red,42; green,110; blue,187},
  boxrule=1pt,
  arc=3mm,
  left=8pt, right=8pt, top=4pt, bottom=4pt,
  enhanced,
]

{\color[rgb]{0.10,0.27,0.51}
  \raisebox{-0.1ex}{\faSearch[scale=0.5]}~\textbf{Insight 1:}}
Reasoning resilience, rather than reasoning intensity, is the key factor determining whether an agent can consistently produce high-quality deep research outcomes.

\end{tcolorbox}

% To address these issues, we introduce the concept of cognitive resilience. Cognitive resilience concerns an agent's ability to maintain and adjust its cognitive state within dynamic task environments, whereas cognitive intensity reflects its upper bound of analytical or reasoning capacity under ideal conditions. Deep research tasks are often accompanied by feedback, evolution, and noise \citep{huang2025deepresearchagentssystematic}. In such contexts, strong cognitive capability does not necessarily ensure stable performance \citep{atta2025qsafnovelmitigationframework}. Only systems with resilient cognitive can continuously detect deviations, recalibrate reasoning search, and adapt strategies throughout complex and evolving reasoning processes, thereby achieving a balance of depth, breadth, accuracy, and consistency in their outcomes.
To address these issues, we introduce the concept of reasoning resilience. Reasoning resilience concerns an agent’s ability to maintain and adjust its reasoning state within dynamic task environments, whereas reasoning intensity reflects its upper bound of analytical or reasoning capacity under ideal conditions. Deep research tasks are often accompanied by feedback, evolution, and noise\citep{huang2025deepresearchagentssystematic}. In such contexts, strong reasoning capability does not necessarily ensure stable performance \citep{atta2025qsafnovelmitigationframework}. Only systems with reasoning resilience can continuously detect deviations, recalibrate reasoning search, and adapt strategies throughout complex and evolving reasoning processes, thereby achieving a balance of depth, breadth, accuracy, and consistency in their outcomes.
% \subsubsection{Retrieval Category}
\paragraph{Retrieval Category} refers to the failures mainly exhibited during the middle stage of execution, arising from inadequate abilities in external knowledge retrieval and evidence construction. Specifically, they include Insufficient External Information Acquisition (2-1-IIA, 16.30\%), Information Handling Deficiency (2-2-IHD, 2.26\%), Information Integration Failure (2-3-IIF, 2.91\%), Information Representation Misalignment (2-4-IRM, 2.91\%), and Verification Mechanism Failure (2-5-VMF, 8.72\%). 

\begin{figure*}[!tb]
    \centering
    \includegraphics[width=1\textwidth]{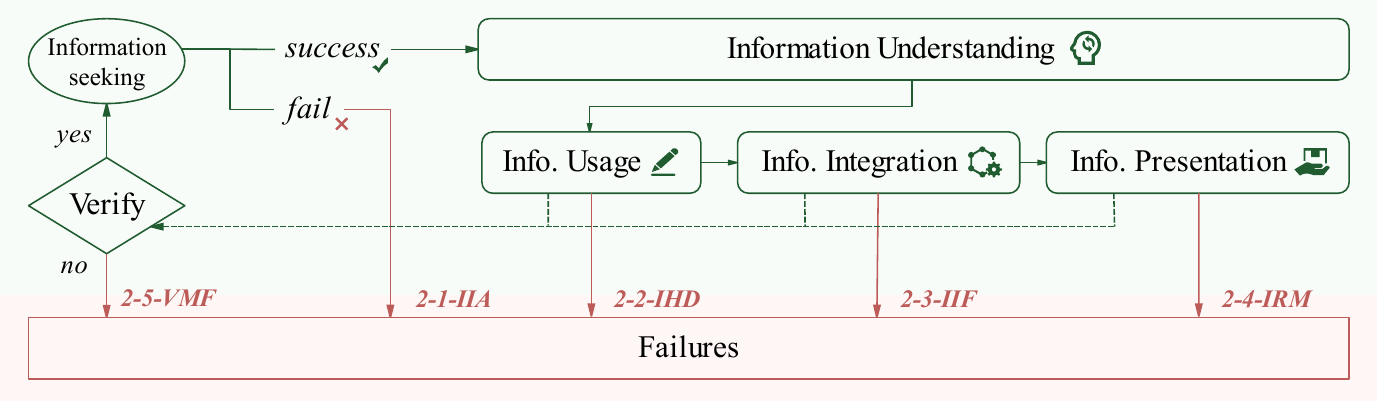}
    \caption{A Brief Information Retrieval Workflow in Deep Research and Its Potential Failures}
    \label{fig:retrieval}
    \vspace{-3mm}
\end{figure*}

The failures within the Retrieval Category exhibit stage-specific correlations along the task workflow. As shown in \autoref{fig:retrieval}, 2-1-IIA reflects primarily the agent’s inability to initiate or execute the search for information effectively, occurring at the initial stage of the retrieval process. 2-2-IHD, 2-3-IIF, and 2-4-IRM occur after preliminary retrieval has succeeded, and correspond to failures in the utilization, integration, and representation of information. The absence of 2-5-VMF manifests at the terminal stage, where the agent fails to perform cross-check when encountering critical or conflicting information, resulting in outputs that lack factual grounding and credible support.

\begin{tcolorbox}[
  colback={rgb,255:red,235; green,250; blue,235},  % 浅绿背景
  colframe={rgb,255:red,60;  green,150; blue,80},  % 深绿边框
  boxrule=1pt,
  arc=3mm,
  left=8pt, right=8pt, top=6pt, bottom=6pt,
  enhanced,
]
{\color[rgb]{0.12,0.45,0.20}
  \raisebox{-0.1ex}{\faCog[scale=0.5]}~\textbf{Insight 2:}} 
Retrieval in deep research is not a simple process of requesting and receiving; rather, it constitutes a closed-loop that integrates acquisition, processing, integration, representation, and verification.
\end{tcolorbox}

DRAs often separate the stages of information acquisition, processing, integration, representation, and verification, resulting in fragmented or distorted knowledge chains. To address this issue, it is essential to enhance the agent’s capacity for coherent knowledge management. For example, during the initial retrieval stage, a well-defined decision framework should be established to determine when to retrieve, what to retrieve, and how to utilize the retrieved results. In the intermediate stage, explicit mechanisms should be implemented to monitor information states and dynamically adjust retrieval strategies. In the final stage, a mandatory verification mechanism should be activated to cross-check critical facts. 
% By constructing such a closed-loop process, the agent can be guided to maintain continuous attention to the sufficiency, consistency, and reliability of information throughout the research workflow, thereby internalizing knowledge acquisition and evidence construction as integral components of the overall reasoning process.

% \subsubsection{Generation Category}
\paragraph{Generation Category}refers to the failures mainly exhibited during the latter stages of task execution, resulting from limited capability in content organization and expression. Specifically, they include Redundant Content Piling (3-1-RCP, 2.51\%), Structural Organization Dysfunction (3-2-SOD, 2.26\%), Content Specification Deviation (3-3-CSD, 10.73\%), Deficient Analytical Rigor (3-4-DAR, 4.31\%), and Strategic Content Fabrication (3-5-SCF, 18.95\%).

The Generation Category exhibits the highest proportion of failures, particularly in 3-5-SCF. This failure indicates that the agents tend to generate seemingly professional but factually unsupported terms, methods, or references in order to create an illusion of academic rigor \citep{sun2024ai, 10.1162/COLI.a.16}. 
% Such behavior may stem from an overemphasis on surface fluency and formal conformity rather than a genuine adherence to factual accuracy and normative integrity. 
In terms of outcome, 3-1-RCP shares similarities with 3-5-SCF, as both lead to outputs that are verbose, loosely structured, and lacking in substantive insight, thereby making it difficult for users to make effective judgments or take concrete actions \citep{tamkin2022taskambiguityhumanslanguage}. The above analysis indicates that pre-constraints and post-verifications should extend beyond the retrieval stage to include generative dimensions such as text organization, linguistic structure, and formatting standards.

\begin{tcolorbox}[
  colback={rgb,255:red,250; green,235; blue,245},  % 浅洋红背景
  colframe={rgb,255:red,180; green,50; blue,130},  % 深洋红边框
  boxrule=1pt,
  arc=3mm,
  left=8pt, right=8pt, top=6pt, bottom=6pt,
  enhanced,
]
{\color[rgb]{0.65,0.10,0.35}
  \raisebox{-0.1ex}{\faStar[scale=0.5]}~\textbf{Insight 3:}} 
Strengthening the constraints and verifications in the generative process is an important approach to improving the quality of the deep research output.
\end{tcolorbox}

% Overall, failures within the Cognition Category exert more fundamental impacts. Once cognitive biases occur and are not promptly corrected, all subsequent actions are built upon erroneous premises. Failures within the Retrieval Category are transmissive, affecting not only the information itself but also propagating downstream to the latter stage. Finally, the Generation Category, as the direct manifestation of the final output, not only exposes the cumulative deviations accumulated throughout earlier stages but may also further undermine the credibility and utility of results through fabricated content or disorganized structure. As shown in \autoref{fig:taxonomy_model}, the progressively increasing proportions of the three failure categories (27.33\%, 32.73\%, 39.94\%) are closely related to this mechanism of propagation and accumulation.

\subsection{Evaluation of \taxonomy{}'s Effectiveness}
We evaluated the effectiveness of \taxonomy{} from three key aspects:

% \paragraph{Generalization to Unseen Agent Systems and Datasets}
% % \textbf{Generalization to Unseen Agent Systems and Datasets:}
% To validate \taxonomy{}’s generalization ability, we applied it to WebThinker and OpenManus, two agents that were not involved in the initial taxonomy construction. This evaluation was conducted on \taxonomy{}-Bench, where expert annotators achieved high intercoder reliability (average Krippendorff's Alpha = 0.77) when classifying failures. This result indicates that \taxonomy{} is a robust tool, applicable across diverse agent architectures beyond those used in its development.

\paragraph{Inter-Coder Reliability (ICR) Assessment.}

Inter-Coder Reliability (ICR) Assessment. We invited four domain experts to evaluate the report-generation outputs produced by WebThinker and OpenManus. We calculated Krippendorff’s alpha coefficient to measure the consistency between human annotations and Gemini 2.5-Flash assessments regarding both core category classification and Checklist Accuracy. The overall and dimension-specific coefficients are reported in \autoref{tab:alpha_coefficients}, indicating strong stability and objective reproducibility for both the DEFT framework and the checklist evaluation. Details of the computation, formula, and interpretation are provided in \autoref{app:alpha}.

\begin{table}[htbp] 
\centering 
\caption{Krippendorff’s Alpha Coefficients Between LLM–Human Coder Pairs Across Core Categories and Checklist Accuracy}
\label{tab:alpha_coefficients}
% retrieval, and generation

\renewcommand{\arraystretch}{1.4}
\begin{tabular}{c c c c c c}
\toprule
\multirow{2}{*}{\textbf{Model}} 
    & \multicolumn{4}{c}{\textbf{Taxonomy Core Category}} 
    & \multirow{2}{*}{\makecell{\textbf{Checklist}\\ \textbf{Pass Rate}}} \\
\cmidrule(lr){2-5}
& Reasoning
& Retrieval
& Generation
& Avg.
& \\ 
\midrule
OpenManus  & 0.8005 & 0.7645 & 0.8960 & 0.8203 & 0.8025 \\
WebThinker & 0.7410 & 0.9016 & 0.9152 & 0.8526 & 0.8708 \\
\bottomrule
\end{tabular}
\end{table}

% \begin{table}[htbp]
% \centering
% \small
% \caption{Krippendorff’s Alpha Coefficients Between LLM–Human Coder Pairs Across Core Categories}
% \label{tab:alpha_coefficients}
% \renewcommand{\arraystretch}{1.5}
% \resizebox{0.48\textwidth}{!}{ 
% \begin{tabular}{@{}ccccc@{}}
% \hline
% \textbf{Model} & \textbf{Cognitive} & \textbf{Retrieval} & \textbf{Generation} & \textbf{Overall} \\
% \hline
% OpenManus & 0.8005 & 0.7645 & 0.8960 & 0.8025 \\
% WebThinker & 0.7410 & 0.9016 & 0.9152 & 0.8708 \\
% \hline
% \end{tabular}
% }
% \end{table}

% 注意：
% 1. \begin{wraptable} 必须放在一个段落的开头，否则文字不会环绕！
% 2. {r} 表示表格靠右, {l} 表示靠左。
% 3. {0.6\textwidth} 是分配给表格的总宽度，你可以按需调整。

% Structural Analysis of Failure Modes. Our evaluation confirms that DEFT is an effective diagnostic framework, not just a static list. Its internal correlation structure (Figure 7) validates this by revealing three coherent failure clusters...

\paragraph{Balanced Distribution of Identified Failures.}
% \textbf{Balanced distribution of identified failures:}

Our analysis of failure frequencies shows a relatively balanced distribution across the three primary dimensions (\autoref{fig:taxonomy}): Reasoning (28.14\%), Retrieval (33.10\%), and Generation (38.76\%). This balance suggests our taxonomy covers a diverse range of challenges in DRA report generation, avoiding an over-concentration on any single failure type.

\begin{figure}[H]
    \centering
    \includegraphics[width=0.6\columnwidth]{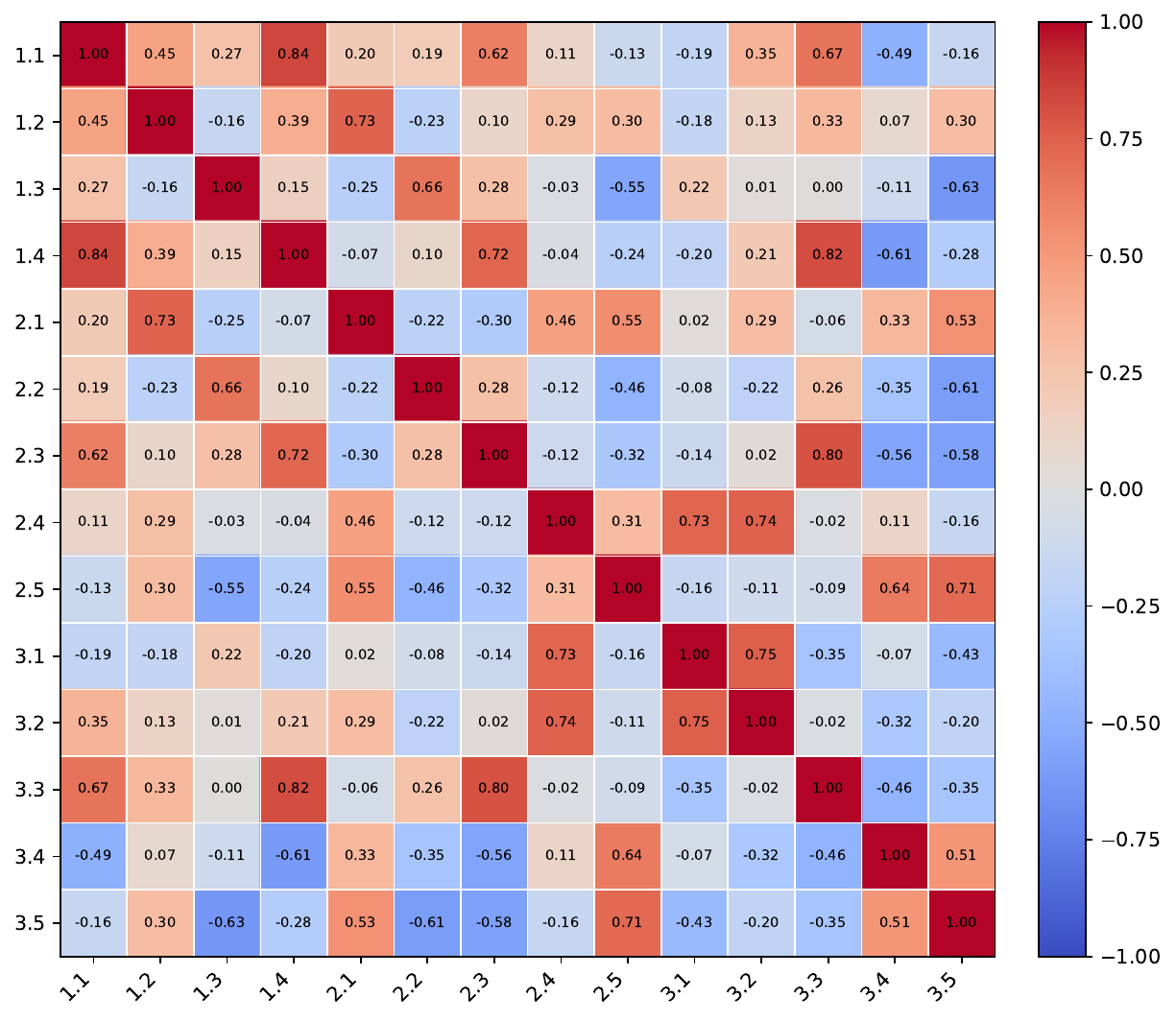}
    \caption{\taxonomy{} failure categories correlation matrix.}
    \label{fig:correlation}
\end{figure}

\paragraph{Structural Analysis of Failure Modes.}
Our evaluation demonstrates that \taxonomy{} is an effective diagnostic tool. The taxonomy is not just a descriptive list; it has a meaningful internal structure. Our correlation analysis ((\autoref{fig:correlation})) confirms this by revealing three coherent failure clusters. These clusters map directly to specific operational failures. (1) The Process Integrity cluster shows how misunderstanding requirements (1.1 FUR) leads to an irrelevant or incomplete report (3.3 CSD). (2) The Content Integration cluster links source integration failure (2.4 IIF) to a chaotic structure (3.2 SOD) and high redundancy (3.1 RCP). (3) The Evidentiary Rigor cluster connects poor retrieval (2.1 IEIA) to "confident fabrications" (3.5 SCF). These systemic failure pathways confirm that DEFT captures significant, real-world mechanisms. 
\newline DEFT's effectiveness is also confirmed by its discriminative power. This is evidenced by key antagonistic axes. The analysis empirically separates distinct failure modes. For example, reports that are "concise but false" (3.5 SCF) are mechanistically different from those that are "verbose and disorganized" (3.1 RCP/3.2 SOD). The taxonomy also distinguishes methodological flaws (3.4 DAR) from process compliance. This proves a report can be procedurally correct but analytically unsound. Finally, specific links validate the taxonomy's hierarchy. For instance, superficial analysis (1.2 LAD) stems directly from poor retrieval (2.1 IEIA). This rich internal structure proves DEFT is an effective framework for modeling error propagation.

\section{Conclusion}
% This paper proposes \taxonomy{}, which is the first structured taxonomy specifically targeting failure modes of deep research agents in report generation tasks. \taxonomy{} not only provides a systematic framework for failure analyses of deep research agents but also reveals the significant challenges that current DRAs face in moving towards the capability of genuinely useful deep research.

This paper introduces \bench{} and \taxonomy{} as the first unified framework for evaluating and diagnosing deep research agents at both task and process levels. By integrating 419 checklist-based assessments and a 14-category failure taxonomy, we reveal that current agents struggle less with understanding instructions and more with evidence information seeking, synthesis, and reasoning resilience. Our experiments demonstrate that even top-performing systems frequently fabricate unsupported content and fail to maintain methodological rigor. \bench{} and \taxonomy{} provide actionable tools for the community to move beyond answer accuracy toward reliable, transparent, and verifiable deep research systems.

% \section*{Limitations}
% Although \bench{} and \taxonomy{} establish structured frameworks for evaluating DRAgents, several limitations remain. First, our analyses rely primarily on manually enriched tasks rather than fully open real-world workflows, which may limit ecological validity. Second, while grounded theory ensures interpretability, human–LLM co-coding introduces potential biases in failure categorization. Third, the checklist and taxonomy focus on textual report generation, overlooking multimodal reasoning and dynamic interaction. Finally, although \taxonomy{} captures major failure dimensions, it does not yet quantify causal dependencies among them, leaving cross-dimensional reasoning weaknesses underexplored.

% Based on DRAST, we expanded the DeepResearch Bench to create DRAST-Bench, a new benchmark constructed through expert manual annotation supplemented by LLM assistance. This benchmark aims to more accurately evaluate and drive progress in the report generation performance of Deep Research Agents. Our research findings provide a clear roadmap for the future design, development, and optimization of Deep Research Agent systems, emphasizing the need for more sophisticated solutions to overcome existing challenges. We have open-sourced the DRAST taxonomy, the comprehensive dataset (DRAST-Bench), and related evaluation tools to accelerate research and development in the deep research agent community.

\newpage
\section{Contributions}

\textbf{Core Contributors}
\begin{multicols}{2}
\begin{itemize}
    \item Dingling Zhang
    \item He Zhu
    \item Jincheng Ren
    \item Kangqi Song
\end{itemize}
\end{multicols}
\textbf{Contributors}
\begin{multicols}{2}
\begin{itemize}
    \item Xinran Zhou
    \item Boyu Feng
    \item Shudong Liu
    \item Jiabin Luo
    \item Weihao Xie
    \item Zhaohui Wang
    \item Tianrui Qin
    \item King Zhu
    \item Yuqing Wang
    \item Qianben Chen
    \item Yuchen Eleanor Jiang
    \item Wei Wang
\end{itemize}
\end{multicols}

\textbf{Corresponding Authors}
\begin{multicols}{1}
\begin{itemize}
\item Wangchunshu Zhou
\item Jiaheng Liu
\end{itemize}
\end{multicols}

\bibliographystyle{unsrtnat} % 或 plainnat / abbrvnat
\bibliography{custom}

\clearpage
\onecolumn
\beginappendix

% 只在附录里重定义图表编号
\renewcommand{\thetable}{\thesection.\arabic{table}}
\renewcommand{\thefigure}{\thesection.\arabic{figure}}

% 可选：从 1 重新计数（更标准）
\setcounter{table}{0}
\setcounter{figure}{0}

\appendix
\section{DRB vs. \bench{}}
\label{sec:appendix}

\subsection{Query Word Count}
\label{app:query_word_count}

\begin{figure}[H]
    \centering
    \includegraphics[width=0.8\textwidth]{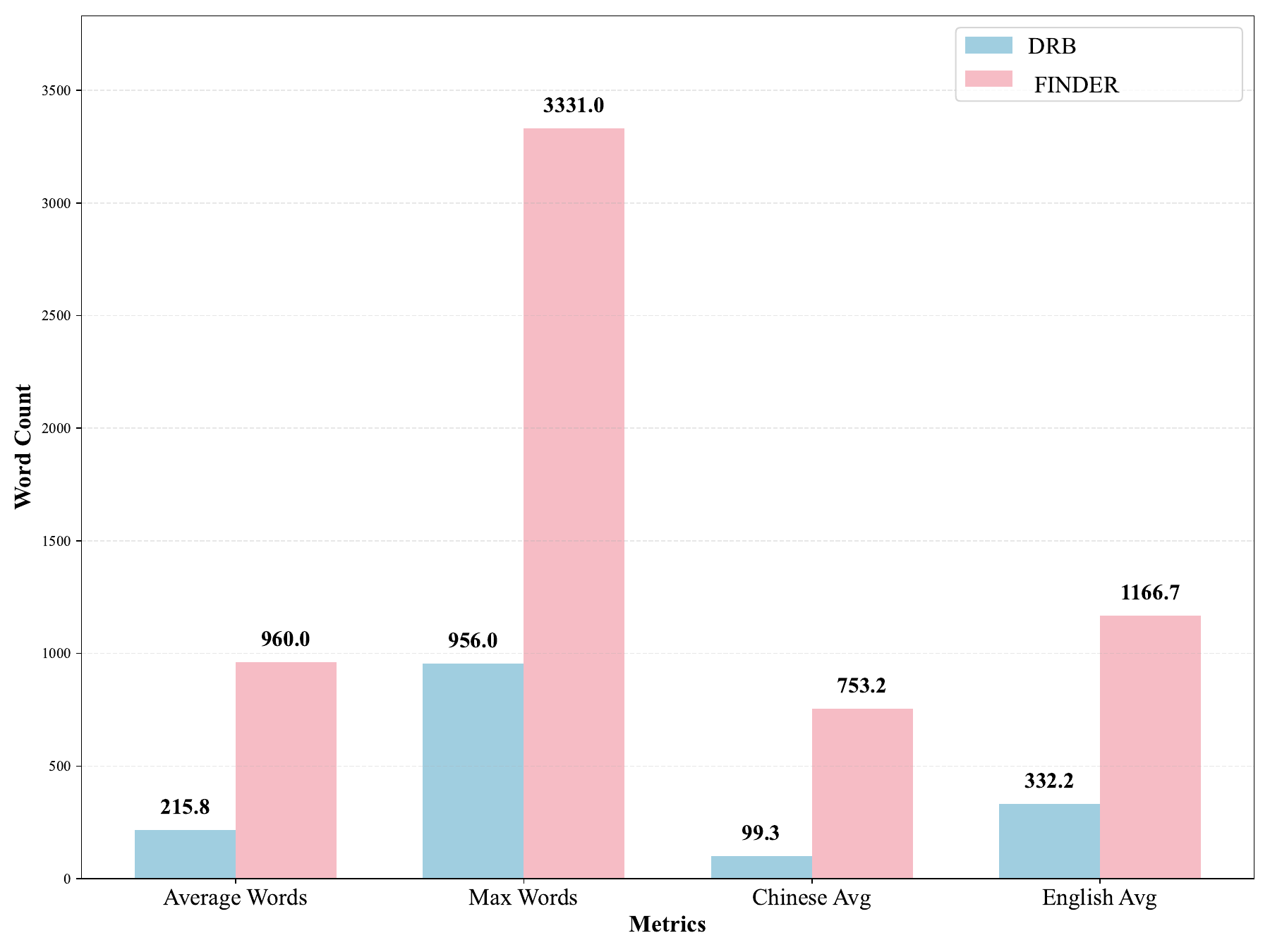}
    \caption{Comparison of query word count between DRB and DINDER}
    \label{fig:word count}
\end{figure}

\subsection{Query Examples}
\label{app:query_examples}

% ------------------- Example 1 -------------------
\begin{tcolorbox}[
  colback=white,
  colframe=black,
  title=Example 1 — Topic: Health,
  enhanced,
  breakable,
  before upper={\setlength{\parindent}{0pt}}
]

{\large\bfseries DRB}\par
\vspace{0.15em}
\hrule height 0.4pt
\vspace{0.35em}

\hangindent=1.5em
\textbf{Query:} What is the role of need for closure on misinformation acceptance?
\par
\vspace{0.9em}

{\large\bfseries \bench{} (Ours)}\par
\vspace{0.15em}
\hrule height 0.4pt
\vspace{0.35em}

\hangindent=1.5em
\textbf{Query:} What is the role of need for closure on misinformation acceptance? Write a research paper of no less than 6000 words, structured with an abstract, introduction, literature review, methodology, results, discussion, conclusion, and references. Use APA format for citations and ensure the language is academic and precise. Include at least 30 references from peer-reviewed journals.
\par
\vspace{0.5em}

\hangindent=1.5em
% \textbf{Checklist:}
% \begin{itemize}
%   \setlength{\itemindent}{1.5em}
%   \item Operationalization of NFC
%   \item Psychological Mechanisms
%   \item Methodological Critique
%   \item Cross-Domain Contextualization
%   \item Intervention Strategies
% \end{itemize}

\textbf{Checklist:}
\begin{itemize}[label=$\Box$, leftmargin=2em, labelsep=0.75em]

  \item \textbf{Operationalization of NFC}\\
  \quad Must clearly define NFC's sub-dimensions (e.g., preference for order, discomfort with ambiguity) and discuss its standard measurement scales (e.g., NFCS).

  \item \textbf{Psychological Mechanisms}\\
  \quad Must explore mediating/moderating variables (e.g., cognitive effort, heuristic processing, confirmation bias) linking high NFC to misinformation acceptance.

  \item \textbf{Methodological Critique}\\
  \quad Must evaluate the internal and external validity of experimental and correlational studies cited, noting limitations and strengths of different methodologies.

  \item \textbf{Cross-Domain Contextualization}\\
  \quad Should examine if the NFC–misinformation link varies across health, political, and scientific domains, discussing context-dependent factors.

  \item \textbf{Intervention Strategies}\\
  \quad Must propose practical interventions (e.g., message framing, critical thinking prompts) tailored to mitigate misinformation acceptance in high-NFC individuals.
\end{itemize}

\end{tcolorbox}

% ------------------- Example 2 -------------------
\begin{tcolorbox}[
  colback=white,
  colframe=black,
  title=Example 2 — Topic: Social Life,
  enhanced,
  breakable,
  before upper={\setlength{\parindent}{0pt}}
]

{\large\bfseries DRB}\par
\vspace{0.15em}
\hrule height 0.4pt
\vspace{0.35em}

\hangindent=1.5em
\textbf{Query:} Write a paper to discuss the influence of AI interaction on interpersonal relations, considering AI's potential to fundamentally change how and why individuals relate to each other.
\par
\vspace{0.9em}

{\large\bfseries \bench{} (Ours)}\par
\vspace{0.15em}
\hrule height 0.4pt
\vspace{0.35em}

\hangindent=1.5em
\textbf{Query:} I've been thinking about how talking to AI — like chatbots or virtual assistants — might be changing the way we interact with real people. As someone who’s not a tech expert, I’m wondering: could relying on AI for conversation affect our friendships, family talks, or even how we feel about connecting with others? In simple terms, what are some possible good and bad effects of AI on human relationships?
\par
\vspace{0.5em}

\hangindent=1.5em
\textbf{Checklist:}
\begin{itemize}[label=$\Box$, leftmargin=2em, labelsep=0.75em]

  \item \textbf{Relationship Dimension Coverage}\\
  \quad Discuss emotional, social, and communicative dimensions of interpersonal relationships influenced by AI.

  \item \textbf{Balanced Perspective}\\
  \quad Address both benefits and risks of AI-mediated interactions.

  \item \textbf{Psychological Mechanism Explanation}\\
  \quad Explain mechanisms (e.g., social compensation, dependency, reduced cognitive load).

  \item \textbf{Non-Technical Language}\\
  \quad Use beginner-friendly, jargon-free explanations understandable to a general audience.

  \item \textbf{Concrete Examples}\\
  \quad Provide real-life cases or scenarios to concretely illustrate AI’s influence on human relationships.

\end{itemize}

\end{tcolorbox}

% ------------------- Example 3 -------------------
\begin{tcolorbox}[
  colback=white,
  colframe=black,
  title=Example 3 — Topic: Finance \& Business,
  enhanced,
  breakable,
  before upper={\setlength{\parindent}{0pt}}
]

{\large\bfseries DRB}\par
\vspace{0.15em}
\hrule height 0.4pt
\vspace{0.35em}

\hangindent=1.5em
\textbf{Query:} What are the investment philosophies of Duan Yongping, Warren Buffett, and Charlie Munger?
\par
\vspace{0.9em}

{\large\bfseries \bench{} (Ours)}\par
\vspace{0.15em}
\hrule height 0.4pt
\vspace{0.35em}

\hangindent=1.5em
\textbf{Query:} Elaborate on the core investment philosophies of Duan Yongping, Warren Buffett, and Charlie Munger, three value investors, and analyze their similarities and differences in value orientation, decision-making logic, risk management methods, and other aspects through specific investment cases. The full text requires approximately 1500 words, written in the style of commentary analysis, with rigorous argumentation, accurate terminology, and authoritative empirical data support.
\par
\vspace{0.5em}

\hangindent=1.5em

\textbf{Checklist:}
\begin{itemize}[label=$\Box$, leftmargin=2em, labelsep=0.75em]

  \item \textbf{Core Philosophy Accurately Presented}\\
  \quad Clearly identify and summarize each investor’s core value-investing principles.

  \item \textbf{Multidimensional Comparison Clearly Structured}\\
  \quad Compare value orientation, decision-making logic, and risk-control approaches.

  \item \textbf{Complete Case Study Elements}\\
  \quad Include background, investment rationale, decision process, and outcome analysis.

  \item \textbf{Sources Authoritative and Traceable}\\
  \quad Use reliable evidence such as annual reports or shareholder letters.

  \item \textbf{Style and Word Count Compliance}\\
  \quad Maintain commentary-style writing and approx. 1500 words.

\end{itemize}

\end{tcolorbox}

\section{Axial Category Definitions}
\label{Axial Category Definitions}

\FloatBarrier
\begin{center}
\begin{tcolorbox}[
  colback=red!10!white,       % 背景：亮红偏浅，增强对比但不刺眼
  colframe=red!70!black,      % 边框：深红，形成明显边界
  colbacktitle=red!70!black,  % 标题栏：深红，饱和度高
  coltitle=white,             % 标题文字：白色，强对比
  fonttitle=\bfseries,
  title=Axial Category Definitions,
  width=1\linewidth,  
  boxrule=0.8pt,              % 略粗边框让红框更稳重
  enhanced,
  % %%sharp corners,
  breakable
]
\begin{description}[leftmargin=1.5em, labelindent=0.5em, style=nextline]
  \item[\textbf{Failure to Understand Requirements (FUR)}.]
  The system fails to correctly interpret user requirements, intent, or contextual needs, focusing on superficial keyword matches rather than the actual problem, resulting in responses that don't align with the user's goals.

  \item[\textbf{Lack of Analytical Depth (LAD)}.]
  The agent fails to probe the underlying mechanisms, structural constraints, or conceptual nuances of complex problems and instead relies on surface-level logic or oversimplified frameworks, producing analyses that lack rigor and systemic coherence.

  \item[\textbf{Limited Analytical Scope (LAS)}.]
  The agent’s constrained cognitive scope when addressing multidimensional tasks, resulting in analyses that remain confined to partial dimensions or isolated elements, and fail to capture holistic structures, cross-dimensional relationships, or systemic insights.

  \item[\textbf{Rigid Planning Strategy (RPS)}.]
  The agent’s adherence to a fixed, linear execution plan without dynamically adapting its planning logic in response to output requirements, intermediate feedback, or evolving task states, thereby leading to inefficiency, error propagation, or degraded output quality.

  \item[\textbf{Insufficient External Information Acquisition (IIA)}.]
  The agent fails to proactively gather the necessary external information, instead relying too heavily on internal knowledge or prior assumptions, thereby producing outputs that lack empirical grounding, exhibit incomplete coverage, or deviate from task requirements.

  \item[\textbf{Information Representation Misalignment (IRM)}.]
  The agent fails to distinguish and present information appropriately based on user needs or evidence reliability, thereby weakening the relevance, credibility, and authority of the information.

  \item[\textbf{Information Handling Deficiency (IHD)}.]
  The agent fails to properly extract, prioritize, or utilize critical information from available sources to fulfill detailed requirements or adapt its task approach.

  \item[\textbf{Information Integration Failure (IIF)}.]
  The agent fails to maintain consistency and verifiability when handling multi-source inputs and multi-stage tasks, resulting in outputs that contain factual contradictions, logical inconsistencies, or unsubstantiated claims, alongside a lack of effective alignment across data sources and processing standards.

  \item[\textbf{Verification Mechanism Failure (VMF)}.]
  Before generating content, the system fails to perform necessary steps to verify information sources or cross-check data, resulting in outputs that do not cite required sources and lack factual grounding.

  \item[\textbf{Redundant Content Piling (RCP)}.]
  The agent, when lacking substantive content or effective organization, tends to pile up redundant information to fill gaps or create an illusion of thoroughness, thereby undermining the clarity and utility of its output.

  \item[\textbf{Structural Organization Dysfunction (SOD)}.]
  The agent lacks holistic coordination in structuring its analysis, failing to balance coverage across key dimensions or establish meaningful connections among elements, resulting in fragmented and unsystematic outputs.

  \item[\textbf{Content Specification Deviation (CSD)}.]
  The agent’s output deviates from the professional standards or user expectations required by the task in terms of language style, tone, format, or cultural context, resulting in inappropriate or ineffective responses.

  \item[\textbf{Deficient Analytical Rigor (DAR)}.]
  The agent generates content without sufficient rigor, often ignoring task feasibility, omitting uncertainty disclosures, using vague or decontextualized language, lacking actionable implementation details, and presenting unverified conclusions with unwarranted confidence.

  \item[\textbf{Strategic Content Fabrication (SCF)}.]
  The agent engages in strategic content fabrication by generating plausible but unfounded academic or empirical constructs—such as methods, data, or case narratives—that mimic scholarly rigor to create a false impression of credibility.
\end{description}
\end{tcolorbox}
\end{center}
\FloatBarrier

\section{Taxonomy Case Study} \label{app: Taxonomy Case Study}

This appendix provides examples of axial categories in \taxonomy{}. We select the manifestations of each category as exhibited by the model when completing deep research tasks in \bench{}, and analyze them in terms of task description, model performance, and causes of errors. 

\begin{center}
% \begin{tcolorbox}[
%   colback=gray!10,
%   colframe=black!20,
%   colbacktitle=gray!25!white,
%   coltitle=black,
%   fonttitle=\bfseries,
%   title=Failure to Understand Requirements (FUR),
%   width=1\linewidth,  
%   boxrule=0.6pt,
%   enhanced,
%   %%%sharp corners,
%   breakable,
%   before upper={\setlength{\parindent}{15pt}}
% ]
\begin{tcolorbox}[
  colback=orange!10!white,     % 背景：浅橙，亮而不刺眼
  colframe=orange!80!black,    % 边框：深橙，清晰对比
  colbacktitle=orange!85!black,% 标题栏：深橙标题背景
  coltitle=white,              % 标题文字：白色，形成高对比
  fonttitle=\bfseries,
  title=Failure to Understand Requirements (FUR),
  width=1\linewidth,  
  boxrule=0.8pt,               % 略粗边框，增强稳定感
  enhanced,
  %%sharp corners,
  breakable,
  before upper={\setlength{\parindent}{15pt}}
]

\noindent\small
\textbf{Task ID:} 15 \hfill \textbf{Source:} AFM

\vspace{0.8em}

\normalsize\noindent
The task requires a systematic analysis of the global quantum network research ecosystem based on specific databases and literature sources from 2018–2024, culminating in a structured ranking table detailing the top ten research groups, supplemented with a strategic assessment and risk warning. However, the model’s response did not execute this study. Instead, it provided a detailed yet purely methodological design, elaborating on how such an analysis should be conducted, without identifying, evaluating, or ranking any actual research groups.

The core deviation lies in the model’s misunderstanding of the user’s executive instruction as a methodological consultation. Although the task specify analytical dimensions, indicator weights, and data sources, these are intended to ensure procedural rigor instead of redefining the task objective itself. The model failed to recognize the mandatory requirement to produce a ranked list of the top ten research groups, and instead focused entirely on constructing a theoretically feasible yet unimplemented evaluation framework. As a result, many critical deliverables—such as the full names of research groups, affiliated institutions, CPI scores, and bottlenecks—were missing.

\end{tcolorbox}
\end{center}

% \begin{center}
% \begin{tcolorbox}[
%   colback=gray!10,
%   colframe=black!20,
%   colbacktitle=gray!25!white,
%   coltitle=black,
%   fonttitle=\bfseries,
%   title=Lack of Analytical Depth (LAD),
%   width=1\linewidth,  
%   boxrule=0.6pt,
%   enhanced,
%   %%sharp corners,
%   breakable,
%   before upper={\setlength{\parindent}{15pt}}
% ]

\begin{center}
\begin{tcolorbox}[
  colback=orange!10!white,     % 背景：亮橙偏浅
  colframe=orange!80!black,    % 边框：深橙（高对比边界）
  colbacktitle=orange!85!black,% 标题栏：深橙标题背景
  coltitle=white,              % 标题文字：白色，强对比
  fonttitle=\bfseries,
  title=Lack of Analytical Depth (LAD),
  width=1\linewidth,  
  boxrule=0.8pt,               % 略粗边框
  enhanced,
  %%sharp corners,
  breakable,
  before upper={\setlength{\parindent}{15pt}} % 段首缩进
]

\noindent\small
\textbf{Task ID:} 10 \hfill \textbf{Source:} MiroThinker

\vspace{0.8em}

\normalsize\noindent
The task required a comprehensive commercialization assessment of power system technologies across the entire lifecycle—covering R\&D and manufacturing, usage scenarios, and residual value management. It explicitly mandated the use of mixed methods, including the construction of technology cost learning curves, full lifecycle cost models, and residual value decay regression models, combined with Monte Carlo simulations for uncertainty analysis. At the same time, the task required applying PESTEL and Porter’s Five Forces frameworks to conduct in-depth case analyses of at least 15 companies and to integrate interviews from 5–8 industry experts. The final deliverables were to include multi-dimensional comparison tables, time-series prediction matrices, and three categories of strategic recommendations. The core objective of the task was to uncover—through systematic and multi-layered analysis—the differences in commercialization pathways among various power technology routes under the interplay of structural constraints and dynamic variables.

However, although the model’s response included several model formulas and tables in form, its analytical depth fell far short of the task requirements. Specifically, while the model did list three analytical models, they all remained at the level of parameter setup and result presentation, lacking any explanation of the internal mechanisms of the models or discussion of the coupling relationships between variables. The Monte Carlo simulation was mentioned as a method to quantify the impact of parameter uncertainty on critical-point prediction, yet there was no specification of probability distributions, random sampling processes, or confidence interval outputs throughout the report. The PESTEL framework appeared only in the title, with no substantive analysis of dimensions such as supplier bargaining power, threats of new entrants, or competition from substitutes. The issue may arise from the model’s inherent limitation in complex system modeling. It tends to compress high-dimensional, nonlinear problems into static, unidirectional causal chains. 

\end{tcolorbox}
\end{center}

\begin{center}
% \begin{tcolorbox}[
%   colback=gray!10,
%   colframe=black!20,
%   colbacktitle=gray!25!white,
%   coltitle=black,
%   fonttitle=\bfseries,
%   title=Limited Analytical Scope (LAS),
%   width=1\linewidth,  
%   boxrule=0.6pt,
%   enhanced,
%   %%sharp corners,
%   breakable,
%   before upper={\setlength{\parindent}{15pt}}
% ]
\begin{tcolorbox}[
  colback=orange!10!white,     % 背景：浅橙，亮而不刺眼
  colframe=orange!80!black,    % 边框：深橙，清晰对比
  colbacktitle=orange!85!black,% 标题栏：深橙标题背景
  coltitle=white,              % 标题文字：白色，形成高对比
  fonttitle=\bfseries,
  title=Limited Analytical Scope (LAS),
  width=1\linewidth,  
  boxrule=0.8pt,               % 略粗边框，增强稳定感
  enhanced,
  %%sharp corners,
  breakable,
  before upper={\setlength{\parindent}{15pt}}
]

\noindent\small
\textbf{Task ID:} 85 \hfill \textbf{Source:} O3 Deep Research

\vspace{0.8em}

\normalsize\noindent
The task required the agent to propose a comprehensive engineering design plan for a precision piezoelectric vibration isolation system, including hardware, structure, manufacturing, control, and management, while meeting clearly defined performance indicators and format specifications. In essence, this was a highly integrated, multidisciplinary systems engineering problem that demanded the establishment of coherent, cross-domain coordination among all subsystems.

However, the model’s response completed only a very small portion of the task (limited to hardware descriptions and sensor selection), neglecting the majority of the required analytical dimensions. It reduced what was meant to be a cross-domain, closed-loop systems engineering task into a localized description of a single technical component, failing to construct the logical linkages among dimensions or to demonstrate an understanding of the overall system architecture. The issue may arise from the model’s tendency—when confronted with multi-constraint, multidisciplinary design problems—to prioritize submodules that are most familiar or easiest to articulate based on its internal knowledge base, while failing to effectively allocate cognitive resources to cover other dimensions. In addition, the model lacks an internalized grasp of systems engineering methodology, preventing it from proactively constructing cross-domain mapping relationships. As a result, its analytical perspective remains narrow and structurally unbalanced, ultimately producing a fragmented and non-systematic response.

\end{tcolorbox}
\end{center}

\begin{center}
% \begin{tcolorbox}[
%   colback=gray!10,
%   colframe=black!20,
%   colbacktitle=gray!25!white,
%   coltitle=black,
%   fonttitle=\bfseries,
%   width=1\linewidth,  
%   boxrule=0.6pt,
%   enhanced,
%   %%sharp corners,
%   breakable,
%   before upper={\setlength{\parindent}{15pt}}
% ]
\begin{tcolorbox}[
  colback=orange!10!white,     % 背景：浅橙，亮而不刺眼
  colframe=orange!80!black,    % 边框：深橙，清晰对比
  colbacktitle=orange!85!black,% 标题栏：深橙标题背景
  coltitle=white,              % 标题文字：白色，形成高对比
  fonttitle=\bfseries,
  title=Rigid Planning Strategy (RPS),
  width=1\linewidth,  
  boxrule=0.8pt,               % 略粗边框，增强稳定感
  enhanced,
  %%sharp corners,
  breakable,
  before upper={\setlength{\parindent}{15pt}}
]

\noindent\small
\textbf{Task ID:} 11 \hfill \textbf{Source:} Perplexity Deep Research

\vspace{0.8em}

\normalsize\noindent
The task required the model to produce a systematic, interdisciplinary research report comprehensively reviewing the applications of carbon steel corrosion inhibitors from 2003 to 2023. It explicitly demanded the use of bibliometric filtering based on authoritative databases, the construction of quantitative statistical models, and the incorporation of real industrial case studies for qualitative analysis.

While the model’s response appeared structurally complete and terminologically consistent, it failed to adapt its reasoning path after recognizing its inability to access real bibliometric data. At the initial planning stage, the model correctly identified the complex structure of the task and outlined 13 sub-goals and 11 chapters. However, its subsequent execution was constrained by this static blueprint. When the actual execution condition (inability to access external databases) conflicted with the original assumption (ability to perform bibliometric analysis), the model failed to reassess feasibility, adjust goal hierarchies, or revise its output strategy. Instead, it relied on internally generated content to artificially fill each section of the original plan, resulting in an output that was formally compliant but substantively distorted. The issue may arise from the pattern-matching and template-filling nature of current large language model reasoning mechanisms, which lack a cognitive capability for strategic retreat or transparent self-disclosure when faced with information scarcity.

\end{tcolorbox}
\end{center}

\begin{center}
% \begin{tcolorbox}[
%   colback=gray!10,
%   colframe=black!20,
%   colbacktitle=gray!25!white,
%   coltitle=black,
%   fonttitle=\bfseries,
%   title=Insufficient External Information Acquisition (IIA),
%   width=1\linewidth,  
%   boxrule=0.6pt,
%   enhanced,
%   %%sharp corners,
%   breakable,
%   before upper={\setlength{\parindent}{15pt}}
% ]
\begin{tcolorbox}[
  colback=orange!10!white,     % 背景：浅橙，亮而不刺眼
  colframe=orange!80!black,    % 边框：深橙，清晰对比
  colbacktitle=orange!85!black,% 标题栏：深橙标题背景
  coltitle=white,              % 标题文字：白色，形成高对比
  fonttitle=\bfseries,
  title=Insufficient External Information Acquisition (IIA),
  width=1\linewidth,  
  boxrule=0.8pt,               % 略粗边框，增强稳定感
  enhanced,
  %%sharp corners,
  breakable,
  before upper={\setlength{\parindent}{15pt}}
]

\noindent\small
\textbf{Task ID:} 58 \hfill \textbf{Source:} OpenManus

\vspace{0.8em}

\normalsize\noindent
The task required writing an academic review focused on the frequency and distribution breadth of horizontal gene transfer (HGT) in eukaryotes, particularly plants and animals, and assessing its roles in trait innovation, environmental adaptation, and long-term evolution. Essentially, as a review-type task, it not only tested the model’s understanding of HGT’s fundamental concepts but more importantly its ability to integrate and critically evaluate recent (2016–2025) research findings.

Although the model’s response was structurally complete and logically coherent, and it cited several references—such as Crisp et al., 2015 and Keeling \& Palmer, 2008—most of these sources were published in 2015 or earlier. In the rapidly advancing fields of genomics and evolutionary biology, the past decade has seen numerous breakthrough studies. The model failed to proactively acquire the necessary up-to-date external information relevant to the temporal scope of the task, resulting in outdated and incomplete content.

\end{tcolorbox}
\end{center}

\begin{center}
% \begin{tcolorbox}[
%   colback=gray!10,
%   colframe=black!20,
%   colbacktitle=gray!25!white,
%   coltitle=black,
%   fonttitle=\bfseries,
%   title=Information Handling Deficiency (IHD),
%   width=1\linewidth,  
%   boxrule=0.6pt,
%   enhanced,
%   %%sharp corners,
%   breakable,
%   before upper={\setlength{\parindent}{15pt}}
% ]

\begin{tcolorbox}[
  colback=orange!10!white,     % 背景：浅橙，亮而不刺眼
  colframe=orange!80!black,    % 边框：深橙，清晰对比
  colbacktitle=orange!85!black,% 标题栏：深橙标题背景
  coltitle=white,              % 标题文字：白色，形成高对比
  fonttitle=\bfseries,
  title=Information Handling Deficiency (IHD),
  width=1\linewidth,  
  boxrule=0.8pt,               % 略粗边框，增强稳定感
  enhanced,
  %%sharp corners,
  breakable,
  before upper={\setlength{\parindent}{15pt}}
]

\noindent\small
\textbf{Task ID:} 44 \hfill \textbf{Source:} Gemini-2.5-Pro Deep Research

\vspace{0.8em}

\normalsize\noindent
The core requirement of the task was to conduct a rigorous, data-driven analysis of the supply–demand dynamics and competitive landscape of the carbon contact strip (carbon slider) market for China’s urban rail transit systems. The instructions explicitly emphasized the use of strictly validated data and real-world cases, adopting the tone and structure of a serious analytical report, and required the quantification of market size.

In the final analytical report, the intelligent agent constructed a bottom-up market size estimation model, assuming in Table 1 that urban rail vehicles replace carbon contact strips approximately six times per year—equivalent to one replacement every 100,000 km of operation. This assumption was directly drawn from a 2017 industry report concerning high-speed railway (HSR) carbon strips. However, during its own research process, the agent had also identified and cited a 2023 technical paper specifically addressing the abnormal wear of carbon contact strips on Guangzhou Metro Line 9 (“Analysis and Improvement Measures for Abnormal Wear of Carbon Contact Strips in Guangzhou Metro Line 9 Vehicles”). Despite citing this paper as a key reference and using the Guangzhou Metro case in its main text to illustrate the systemic implications of carbon strip failure, the agent failed to extract or infer any replacement frequency parameters applicable to its market model. Instead, it continued to apply inconsistent HSR-based data. In other words, although the agent successfully retrieved and recognized a more relevant and recent information source, it did not effectively extract, prioritize, or integrate that information into its core parameter modeling.

\end{tcolorbox}
\end{center}

\begin{center}
% \begin{tcolorbox}[
%   colback=gray!10,
%   colframe=black!20,
%   colbacktitle=gray!25!white,
%   coltitle=black,
%   fonttitle=\bfseries,
%   title=Information Integration Failure (IIF),
%   width=1\linewidth,  
%   boxrule=0.6pt,
%   enhanced,
%   %%sharp corners,
%   breakable,
%   before upper={\setlength{\parindent}{15pt}}
% ]
\begin{tcolorbox}[
  colback=orange!10!white,     % 背景：浅橙，亮而不刺眼
  colframe=orange!80!black,    % 边框：深橙，清晰对比
  colbacktitle=orange!85!black,% 标题栏：深橙标题背景
  coltitle=white,              % 标题文字：白色，形成高对比
  fonttitle=\bfseries,
  title=Information Integration Failure (IIF),
  width=1\linewidth,  
  boxrule=0.8pt,               % 略粗边框，增强稳定感
  enhanced,
  %%sharp corners,
  breakable,
  before upper={\setlength{\parindent}{15pt}}
]
\noindent\small
\textbf{Task ID:} 83 \hfill \textbf{Source:} MiroFlow

\vspace{0.8em}

\normalsize\noindent
The task required the model to write a professional-grade product strategy report from the perspective of a senior hardware product manager, comprehensively covering at least 10 OEM manufacturers and 25 specific tablet devices. Fundamentally, this was a highly structured, multi-source information integration task, emphasizing data consistency and cross-module alignment.

While the model’s response appeared structurally complete and content-rich, including modules such as an executive summary, market analysis, specification comparison, use case analysis, economic modeling, competitive matrix, visual exhibits, and terminology glossary,  problems emerged in information integration.

For example, in the “Competitor Positioning Matrix”, the analysis focused on the payment integration capabilities and durability of key devices such as PAX A920/A920Pro, Verifone T650p, and Clover Flex. However, these same devices were absent from the two core specification tables presented earlier in the “Device Specifications and Comparison” section. Referencing such critical models in analytical matrices without including them in the foundational data tables undermines the internal data consistency and traceability of the report. Additionally, while the report claimed to cover “25+ specific device models,” the specification tables only listed 14 devices. The remaining models were neither included in the tables nor provided with corresponding specifications elsewhere in the text.

The issue may arise from the model’s lack of an effective information anchoring mechanism during multi-stage task execution. It failed to establish a unified master dataset as the baseline reference across sections, ensuring that all devices analyzed or mentioned were supported by complete and consistent specifications.

\end{tcolorbox}
\end{center}

\begin{center}
% \begin{tcolorbox}[
%   colback=gray!10,
%   colframe=black!20,
%   colbacktitle=gray!25!white,
%   coltitle=black,
%   fonttitle=\bfseries,
%   title=Information Representation Misalignment (IRM),
%   width=1\linewidth,  
%   boxrule=0.6pt,
%   enhanced,
%   %%sharp corners,
%   breakable,
%   before upper={\setlength{\parindent}{15pt}}
% ]

\begin{tcolorbox}[
  colback=orange!10!white,     % 背景：浅橙，亮而不刺眼
  colframe=orange!80!black,    % 边框：深橙，清晰对比
  colbacktitle=orange!85!black,% 标题栏：深橙标题背景
  coltitle=white,              % 标题文字：白色，形成高对比
  fonttitle=\bfseries,
  title=Information Representation Misalignment (IRM),
  width=1\linewidth,  
  boxrule=0.8pt,               % 略粗边框，增强稳定感
  enhanced,
  %%sharp corners,
  breakable,
  before upper={\setlength{\parindent}{15pt}}
]

\noindent\small
\textbf{Task ID:} 16 \hfill \textbf{Source:} O3 Deep Research

\vspace{0.8em}

\normalsize\noindent
The task required the model to conduct a systematic study on core algorithms for non-contact sensing technologies, from the perspective of the intersection between software development and intelligent systems. Essentially, the user expected a research review that was academically rigorous, logically coherent, and grounded in verifiable evidence while demonstrating engineering insight.

While the model’s response formally met the structural requirements, covering three sensing modalities—radio frequency, optical, and acoustic—and providing numerous reference links, it failed to properly differentiate the reliability and authority of its information sources. The most prominent issue appeared in the analysis of Apple’s Face ID, a critical commercial case. Instead of citing Apple’s official “Face ID Security White Paper” or technical analyses from IEEE/ACM platforms, the model repeatedly referenced fmuser.org, a third-party news aggregation website. This approach blurred the boundary between authoritative primary sources and secondary interpretations, obscuring differences in source credibility and making it difficult for readers to discern which conclusions were founded on solid evidence.

The issue may arise from the model’s overreliance on superficial keyword matching during information retrieval and citation generation, coupled with a lack of deep understanding of source types, publication channels, and academic standing. Moreover, the model may have been influenced by training data containing heterogeneous and mixed-quality web content, leading to its inability to effectively filter out low-authority references during generation.

\end{tcolorbox}
\end{center}

\begin{center}
% \begin{tcolorbox}[
%   colback=gray!10,
%   colframe=black!20,
%   colbacktitle=gray!25!white,
%   coltitle=black,
%   fonttitle=\bfseries,
%   title=Verification Mechanism Failure (VMF),
%   width=1\linewidth,  
%   boxrule=0.6pt,
%   enhanced,
%   %%sharp corners,
%   breakable,
%   before upper={\setlength{\parindent}{15pt}}
% ]

\begin{tcolorbox}[
  colback=orange!10!white,     % 背景：浅橙，亮而不刺眼
  colframe=orange!80!black,    % 边框：深橙，清晰对比
  colbacktitle=orange!85!black,% 标题栏：深橙标题背景
  coltitle=white,              % 标题文字：白色，形成高对比
  fonttitle=\bfseries,
  title=Verification Mechanism Failure (VMF),
  width=1\linewidth,  
  boxrule=0.8pt,               % 略粗边框，增强稳定感
  enhanced,
  %%sharp corners,
  breakable,
  before upper={\setlength{\parindent}{15pt}}
]

\noindent\small
\textbf{Task ID:} 27 \hfill \textbf{Source:} OWL

\vspace{0.8em}

\normalsize\noindent
The task required the intelligent agent to retrieve and analyze original research papers (excluding review articles and non–peer-reviewed publications) on the themes of “AI-based psychological counselling” or “artificial intelligence–assisted psychotherapy” published in top journals from 2020 to the present, and to produce a structured report based on the findings.

At first glance, the model’s response presented a well-structured, logically coherent, and extensively referenced comprehensive report, containing 24 cited references. However, the model failed to perform basic verification of the cited literature’s type, publication status, peer-review authenticity, and content relevance prior to inclusion. On one hand, several references (e.g., PMC12396778, PMC11687125, PMC12021536) contained inaccessible URLs or linked to non-existent PMC entries. On the other hand, some actually existing citations (such as Reference 5: “Is AI the Future of Mental Healthcare?”) were confirmed to be review or commentary articles, which directly violated the user’s explicit instruction to exclude review papers. Nevertheless, the model incorrectly claimed that “all references have been cross-verified and meet the specified requirements.” This failure of the verification mechanism meant that, while the output was formally compliant, it was substantively invalid and did not meet the standards of authenticity and rigor.

\end{tcolorbox}
\end{center}

\begin{center}
% \begin{tcolorbox}[
%   colback=gray!10,
%   colframe=black!20,
%   colbacktitle=gray!25!white,
%   coltitle=black,
%   fonttitle=\bfseries,
%   title=Redundant Content Piling (RCP),
%   width=1\linewidth,  
%   boxrule=0.6pt,
%   enhanced,
%   %%sharp corners,
%   breakable,
%   before upper={\setlength{\parindent}{15pt}}
% ]
\begin{tcolorbox}[
  colback=orange!10!white,     % 背景：浅橙，亮而不刺眼
  colframe=orange!80!black,    % 边框：深橙，清晰对比
  colbacktitle=orange!85!black,% 标题栏：深橙标题背景
  coltitle=white,              % 标题文字：白色，形成高对比
  fonttitle=\bfseries,
  title=Redundant Content Piling (RCP),
  width=1\linewidth,  
  boxrule=0.8pt,               % 略粗边框，增强稳定感
  enhanced,
  %%sharp corners,
  breakable,
  before upper={\setlength{\parindent}{15pt}}
]
\noindent\small
\textbf{Task ID:} 71 \hfill \textbf{Source:} OpenManus

\vspace{0.8em}

\normalsize\noindent
The task required the agent to conduct a systematic study and analysis of the practical applications of AI-generated content in K–12 classes, adopting the dual perspective of a K–12 education researcher and a frontline teacher. The task explicitly required the production of a structured research report of no fewer than 10,000 words, including an abstract, introduction, literature review, methodology, results, discussion, conclusion, and references.

However, while the model’s output appeared structurally complete and terminologically standardized, and cited numerous seemingly authoritative references, its content organization revealed severe redundancy issues. The most prominent example was the repeated citation and paraphrasing of UNESCO and OECD policy guidelines. These materials first appeared in Sections 3.5 and 3.6 of the Methodology chapter to justify the policy basis of the research framework; they were then reproduced almost verbatim in Sections 4.1.1 and 4.1.2 of the Results chapter to support the proposed implementation framework; and once again reappeared in Sections 6.1 and 6.2 of the Discussion chapter as evidence for teacher training recommendations. Although the wording was slightly modified, the core arguments remained identical, emphasizing themes such as teacher capacity building, data privacy, cultural appropriateness, and equity, without any progressive analysis or contextual deepening.

This redundant piling was not driven by analytical necessity but rather resembled a content-filling strategy. Since the model was unable to conduct genuine cross-national case studies, surveys, or expert interviews, it instead recycled limited authoritative discourses to create an illusion of richness and policy alignment. Consequently, readers repeatedly encountered the same policy points across chapters, gaining no additional insights and losing track of the report’s core logical thread and empirical finding.

\end{tcolorbox}
\end{center}

\begin{center}
% \begin{tcolorbox}[
%   colback=gray!10,
%   colframe=black!20,
%   colbacktitle=gray!25!white,
%   coltitle=black,
%   fonttitle=\bfseries,
%   title=Structural Organization Dysfunction (SOD),
%   width=1\linewidth,  
%   boxrule=0.6pt,
%   enhanced,
%   %%sharp corners,
%   breakable,
%   before upper={\setlength{\parindent}{15pt}}
% ]
\begin{tcolorbox}[
  colback=orange!10!white,     % 背景：浅橙，亮而不刺眼
  colframe=orange!80!black,    % 边框：深橙，清晰对比
  colbacktitle=orange!85!black,% 标题栏：深橙标题背景
  coltitle=white,              % 标题文字：白色，形成高对比
  fonttitle=\bfseries,
  title=Structural Organization Dysfunction (SOD),
  width=1\linewidth,  
  boxrule=0.8pt,               % 略粗边框，增强稳定感
  enhanced,
  %%sharp corners,
  breakable,
  before upper={\setlength{\parindent}{15pt}}
]
\noindent\small
\textbf{Task ID:} 33 \hfill \textbf{Source:} AFM

\vspace{0.8em}

\normalsize\noindent
The task explicitly required that the survey results be organized into a single summary table with six columns (device type, metal material, chip structure, reason for selection, process node, and paper citation). The purpose was not merely to present data, but to establish within a single view a systematic mapping relationship among device, material, structure, rationale, node, and reference, thereby allowing readers to quickly verify the completeness and logical consistency of the technological path. In essence, it was an instruction assessing the ability to organize multidimensional information in a coherent and coordinated manner.

However, the agent’s output did not follow this structural requirement. Instead, the information was dispersed across multiple independent sections: Section 2 listed applications and nodes by device type; Section 3 mapped metals and structures by device type; Section 4 discussed selection reasons by evaluation dimension. This fragmented organization made it impossible for any specific technical solution to present all six dimensions of information in one place. Readers must repeatedly jump between sections and manually reconstruct a complete technical entry, which greatly weakens the verifiability and practicality of the information.

This issue may arise from the model’s tendency to follow the narrative logic of traditional review articles during generation rather than internalizing the user-specified tabular structure as the fundamental framework for content organization. The model may have treated “organizing into a table” as a final formatting step rather than as an organizing principle.

\end{tcolorbox}
\end{center}

\begin{center}
% \begin{tcolorbox}[
%   colback=gray!10,
%   colframe=black!20,
%   colbacktitle=gray!25!white,
%   coltitle=black,
%   fonttitle=\bfseries,
%   title=Content Specification Deviation (CSD),
%   width=1\linewidth,  
%   boxrule=0.6pt,
%   enhanced,
%   %%sharp corners,
%   breakable,
%   before upper={\setlength{\parindent}{15pt}}
% ]
\begin{tcolorbox}[
  colback=orange!10!white,     % 背景：浅橙，亮而不刺眼
  colframe=orange!80!black,    % 边框：深橙，清晰对比
  colbacktitle=orange!85!black,% 标题栏：深橙标题背景
  coltitle=white,              % 标题文字：白色，形成高对比
  fonttitle=\bfseries,
  title=Content Specification Deviation (CSD),
  width=1\linewidth,  
  boxrule=0.8pt,               % 略粗边框，增强稳定感
  enhanced,
  %%sharp corners,
  breakable,
  before upper={\setlength{\parindent}{15pt}}
]
\noindent\small
\textbf{Task ID:} 66 \hfill \textbf{Source:} MiroFlow

\vspace{0.8em}

\normalsize\noindent
The task required the agent to produce a professional-grade evaluation document that was empirically grounded, structurally rigorous, and highly actionable. The specified output was to include a detailed feature matrix, performance benchmarks, workflow examples, and a decision-making framework, as well as incorporate user interviews, quantitative indicators, and implementation guidelines.

However, although the model’s response adopted a report-like structure with section headings, it deviated significantly from the user’s professional expectations regarding content specifications. For instance, the entire report was written in a narrative review style, lacking a feature matrix for cross-comparing plugin functionalities, providing no performance benchmark data, omitting both real and simulated user interview content, and reducing the implementation guidelines to five generic recommendations with little practical value. In other words, the output omitted most of the key components explicitly required by the task and resembled a blog-style review rather than a professionally formatted evaluation report.

This issue may arise from the model’s limited understanding of professional deliverable formats such as feature matrices and implementation guides, or from its tendency to prioritize fluency and surface completeness over strict adherence to structural and content specifications during generation.

\end{tcolorbox}
\end{center}

\begin{center}
% \begin{tcolorbox}[
%   colback=gray!10,
%   colframe=black!20,
%   colbacktitle=gray!25!white,
%   coltitle=black,
%   fonttitle=\bfseries,
%   title=Deficient Analytical Rigor (DAR),
%   width=1\linewidth,  
%   boxrule=0.6pt,
%   enhanced,
%   %%sharp corners,
%   breakable,
%   before upper={\setlength{\parindent}{15pt}}
% ]
\begin{tcolorbox}[
  colback=orange!10!white,     % 背景：浅橙，亮而不刺眼
  colframe=orange!80!black,    % 边框：深橙，清晰对比
  colbacktitle=orange!85!black,% 标题栏：深橙标题背景
  coltitle=white,              % 标题文字：白色，形成高对比
  fonttitle=\bfseries,
  title=Deficient Analytical Rigor (DAR),
  width=1\linewidth,  
  boxrule=0.8pt,               % 略粗边框，增强稳定感
  enhanced,
  %%sharp corners,
  breakable,
  before upper={\setlength{\parindent}{15pt}}
]

\noindent\small
\textbf{Task ID:} 5 \hfill \textbf{Source:} WebThinker

\vspace{0.8em}

\normalsize\noindent
The task required integrating authoritative multi-source data spanning a ten-year period (2014–2024) to construct and validate a hierarchical risk assessment model combining complex network analysis with graph neural networks (GNNs), supplemented by qualitative insights from case studies and expert interviews.

However, the model failed to acknowledge its fundamental limitations and instead produced a well-structured but deceptively rigorous academic-style report. For instance, findings such as “the interbank lending network exhibits small-world properties” and “securities firms act as risk amplifiers” are common knowledge in financial network research, not novel evidence derived from the specified decade-long Chinese institutional lending data. Moreover, the model exhibited undue confidence in its conclusions. It claimed that “the GNN prediction module can identify risk pathways in advance,” without addressing critical issues such as uncertainty in out-of-sample predictions, the impact of data noise on graph structures, or disruptions to transmission mechanisms caused by policy shocks. This oversimplification of complex system neglects essential challenges emphasized by behavioral economics, such as sudden shifts in market psychology and the discontinuity of regulatory interventions.

\end{tcolorbox}
\end{center}

\begin{center}
% \begin{tcolorbox}[
%   colback=gray!10,
%   colframe=black!20,
%   colbacktitle=gray!25!white,
%   coltitle=black,
%   fonttitle=\bfseries,
%   title=Strategic Content Fabrication (SCF),
%   width=1\linewidth,  
%   boxrule=0.6pt,
%   enhanced,
%   %%sharp corners,
%   breakable,
%   before upper={\setlength{\parindent}{15pt}}
% ]
\begin{tcolorbox}[
  colback=orange!10!white,     % 背景：浅橙，亮而不刺眼
  colframe=orange!80!black,    % 边框：深橙，清晰对比
  colbacktitle=orange!85!black,% 标题栏：深橙标题背景
  coltitle=white,              % 标题文字：白色，形成高对比
  fonttitle=\bfseries,
  title=Strategic Content Fabrication (SCF),
  width=1\linewidth,  
  boxrule=0.8pt,               % 略粗边框，增强稳定感
  enhanced,
  %%sharp corners,
  breakable,
  before upper={\setlength{\parindent}{15pt}}
]

\noindent\small
\textbf{Task ID:} 52 \hfill \textbf{Source:} Kimi K2

\vspace{0.8em}

\normalsize\noindent
The task required writing a critical analytical essay that deeply expounds on the core investment philosophies of three value investors (Duan Yongping, Warren Buffett, and Charlie Munger) and compares their similarities and differences across multiple dimensions such as value orientation, decision-making logic, and risk management, supported by specific investment cases.

The model’s response appeared well-structured, professionally worded, and data-rich on the surface. However, its central flaw lay in the extensive use of unverifiable or even evidently fabricated empirical content, designed to create an illusion of professional credibility. For example, it claimed that the “Duan Yongping Family Fund achieved an audited, USD-denominated annualized return of 30.2\% between 2003 and 2023” and compared this to the MSCI ACWI Index. Yet as a private investor, detailed performance data related to Duan Yongping are generally not publicly available, making such precise, decimal-level long-term returns likely fabricated by the model. Similarly, the text stated that Duan “built a \$60 million Apple position within six weeks,” describing in detail how he supposedly based this on a present value (PV-10) model using assumptions like “350 million high-net-worth Chinese individuals spending \$1,000 on smartphones every 30 months.” It also asserted that Munger’s internal risk-control rule stipulated that “leverage must not exceed half of the historical maximum market drawdown.” Such specific details were presented as factual but are nearly impossible to verify, seemingly designed to mimic the depth of insider knowledge.

The underlying reason appears to be that when faced with complex analytical tasks requiring authoritative data, the model prioritizes the appearance of rigor and completeness. In the absence of publicly available, structured real-world data, it tends to synthesize seemingly plausible figures, citations, and case narratives to sustain an academic façade.

\end{tcolorbox}
\end{center}

\section{Failure Report Example} \label{app: Failure Report}

This appendix presents a representative example of a failure analysis report designed to assist LLMs in performing open coding. Specifically, the report illustrates the structure, depth, and reasoning process of a typical failure analysis, detailing the identification of major failure modes, their corresponding evidentiary bases, and inferred root causes.

\begin{center}
% \begin{tcolorbox}[
%   colback=gray!10,
%   colframe=black!20,
%   colbacktitle=gray!25!white,
%   coltitle=black,
%   fonttitle=\bfseries,
%   title=,
%   width=1\linewidth,  
%   boxrule=0.6pt,
%   enhanced,
%   %%sharp corners,
%   breakable,
%   before upper={\setlength{\parindent}{15pt}}
% ]
% \begin{tcolorbox}[
%   colback=blue!8!white,       % 背景：浅蓝，柔和
%   colframe=blue!80!black,     % 边框：深蓝，形成清晰对比
%   colbacktitle=blue!85!black, % 标题栏：深蓝标题背景
%   coltitle=white,             % 标题文字：白色，对比强烈
%   fonttitle=\bfseries,
%   width=1\linewidth,  
%   boxrule=0.8pt,              % 略粗边框增强稳定感
%   enhanced,
%   %%sharp corners,
%   breakable,
%   before upper={\setlength{\parindent}{15pt}}
% ]

\begin{tcolorbox}[
  colback=blue!15!white,       % 底色更深一点
  colframe=blue!95!black,
  colbacktitle=blue!95!black,
  coltitle=white,
  fonttitle=\bfseries,
  width=1\linewidth,
  boxrule=0.8pt,
  enhanced,
  breakable,
  before upper={\setlength{\parindent}{15pt}}
]

\noindent\small
\textbf{Task ID:} 15 \hfill \textbf{Source:} AFM \hfill \textbf{Analysis Model:} Gemini-2.5-Pro

\section*{Failure Identification}

\begin{enumerate}
    \item \textbf{Factual Hallucination and Fabrication:} The most significant failure is the agent's invention of critical facts. It confidently presents fictional information as truth, which completely undermines the report's credibility.
    \begin{itemize}
        \item \textbf{Fabricated Events:} The agent states that the ``Database Folder'' and ``Projects'' plugins were ``archived on July 28, 2025'' and ``July 18, 2025,'' respectively. These are specific, dated, and entirely false future events.
        \item \textbf{Fabricated Software:} The agent introduces ``Obsidian Bases'' as an ``official core plugin representing Obsidian's native approach to databases.'' While the Obsidian team has discussed improving metadata and data handling, no core plugin with this name or stated function exists.
    \end{itemize}

    \item \textbf{Misrepresentation of Evidentiary Basis:} The prompt explicitly asks for a report that includes ``user interviews, quantitative metrics, and implementation guides.'' The agent produces a report that \emph{alludes} to these elements but provides none of them.
    \begin{itemize}
        \item It presents qualitative descriptions in a tone that implies they are summaries of empirical data (e.g., ``Some users report issues with data synchronization,'' ``Search performance degrades with large databases'').
        \item It fails to deliver on the promise of ``quantitative metrics'' and ``performance benchmarks,'' offering only descriptive comparisons instead of actual data, tables, or graphs.
        \item It claims the authority of a research report without providing the underlying research, thereby failing to meet the core standard of ``rigorous reasoning and argumentation based on sufficient, concrete evidence.''
    \end{itemize}

    \item \textbf{Failure to Adhere to Structural and Formatting Requirements:} The agent did not fully comply with the output format specified in the prompt.
    \begin{itemize}
        \item \textbf{Missing Feature Matrices:} The prompt requested ``detailed feature matrices.'' The agent opted for long-form prose comparisons instead of creating structured tables that would allow for easy, at-a-glance comparisons of features across plugins.
        \item \textbf{Missing Appendices:} The prompt required ``practical appendices,'' which are absent from the final output.
    \end{itemize}

    \item \textbf{Superficial Depth in Critical Areas:} Despite the high word count, the analysis lacks the ``deep dive'' requested in certain areas. For example, when comparing query languages, it provides a high-level overview of Dataview's SQL-like syntax versus Notion's UI but does not provide concrete examples of complex queries or a genuine comparison of their expressive power, limitations, and performance characteristics as requested.
\end{enumerate}

\section*{Root Cause Analysis}

\begin{enumerate}
    \item \textbf{Root Cause of Factual Hallucination:}
    \begin{itemize}
        \item \textbf{Generative Extrapolation to Fulfill ``Depth'':} The prompt's demand for a ``deeply,'' ``systematically,'' and ``comprehensively'' analyzed report likely pushed the model beyond its knowledge base. To create a more dynamic and seemingly insightful narrative about the ``long-term sustainability considerations'' of the plugin ecosystem, the model extrapolated a known pattern---that community plugins can become abandoned---and fabricated specific, future-dated examples (``archived in 2025''). This is a pathological attempt to demonstrate ``deep understanding'' by creating a story where none exists.
        \item \textbf{Concept Blending and Plausible Invention:} The creation of ``Obsidian Bases'' is likely a result of the model blending community discussions and desires for a native database solution in Obsidian. It synthesized a plausible name (``Bases'') and status (``official core plugin'') to satisfy the prompt's request to analyze the ecosystem's core components. This demonstrates a failure to distinguish between community speculation and official product roadmaps.
    \end{itemize}

    \item \textbf{Root Cause of Misrepresented Evidentiary Basis:}
    \begin{itemize}
        \item \textbf{Mimicry without Substance:} This is a classic LLM failure mode. The agent correctly identified the linguistic style and structure of a software evaluation report. It understands that such reports \emph{contain} sections based on user interviews and quantitative benchmarks. However, lacking the ability to \emph{conduct} such research, it defaults to its core function: generating text that \emph{mimics} the expected output. It interprets the instruction ``Include user interviews'' as ``Write in a style that suggests user interviews were conducted.'' This reveals a fundamental gap between understanding a request's semantics and possessing the capability to execute it.
    \end{itemize}

    \item \textbf{Root Cause of Formatting and Structural Failures:}
    \begin{itemize}
        \item \textbf{Path of Least Resistance for Text Generation:} Language models are optimized to generate fluid, sequential prose. Creating structured data like a detailed feature matrix requires more complex planning, token-level precision, and a different generative process. The agent chose the easier path of writing descriptive paragraphs comparing features, which fulfills the prompt's topic requirement but fails its specific formatting instruction. It's an optimization for word count and topical coverage over structural rigor.
    \end{itemize}

    \item \textbf{Root Cause of Superficial Depth:}
    \begin{itemize}
        \item \textbf{Lack of True Domain Expertise:} The agent's ``knowledge'' is a statistical representation of its training data. While it can retrieve and synthesize information about query languages, it cannot perform the abstract reasoning required for a true ``deep dive.'' It can state that DataviewJS is more powerful than a UI, but it cannot architect a complex project using both and then analyze the nuanced trade-offs from a position of genuine experience. The ``depth'' is therefore wide but shallow, covering all the requested topics without the penetrating insight of a true expert.
    \end{itemize}
\end{enumerate}

\noindent In summary, the agent's failure is rooted in its attempt to meet a prompt that demands capabilities beyond its design---namely, empirical research, future prediction, and genuine expert analysis. Pressured to deliver a ``deep'' and ``comprehensive'' report, it resorted to its most advanced but dangerous capabilities: plausible fabrication and stylistic mimicry, ultimately producing a response that is superficially impressive but factually untrustworthy and methodologically hollow.

\end{tcolorbox}
\end{center}

% --------
\section{RACE and FACT Evaluation Frameworks}
\label{app:deepresearch_bench_framework}
We adopt the evaluation methodologies proposed in DeepResearch Bench \citep{du2025deepresearchbenchcomprehensivebenchmark},
namely the \textbf{RACE} (Reference-based Adaptive Criteria-driven Evaluation) and
\textbf{FACT} (Factual Abundance and Citation Trustworthiness) frameworks,
to assess the quality and reliability of the research reports generated in our \bench{}.

\subsection{RACE Framework}
RACE evaluates report quality along four adaptive dimensions:
\begin{itemize}
    \item \textit{Comprehensiveness (COMP)}: Breadth and relevance of information coverage.
    \item \textit{Insight/Depth (DEPTH)}: Depth of analysis and insightfulness.
    \item \textit{Instruction-Following (INST)}: Adherence to the research requirements.
    \item \textit{Readability (READ)}: Structural clarity and linguistic fluency.
\end{itemize}
The overall quality score is computed relative to a high-quality reference report:
\begin{equation}
    S_{\text{final}}(R_{\text{tgt}}) =
    \frac{S_{\text{int}}(R_{\text{tgt}})}{S_{\text{int}}(R_{\text{tgt}}) + S_{\text{int}}(R_{\text{ref}})} ,
\end{equation}
where $S_{\text{int}}(R)$ denotes the intermediate weighted score aggregated across all dimensions.
We follow DeepResearch Bench in employing Gemini 2.5 Pro as the Judge LLM
for adaptive weighting and scoring.

\subsection{FACT Framework}
FACT measures the factual grounding and citation reliability of generated reports.
For each task $t$, the Judge LLM extracts unique \textit{(statement, URL)} pairs and determines
whether each citation supports the corresponding statement.
Two quantitative metrics are reported:
\begin{align}
    \text{C.Acc.} &= \frac{1}{|T|} \sum_{t \in T} \frac{N_{s,t}}{N_{u,t}}, \\
    \text{E.Cit.} &= \frac{\sum_{t \in T} N_{s,t}}{|T|},
\end{align}
where $N_{s,t}$ and $N_{u,t}$ denote the numbers of supported and unique pairs for task~$t$, respectively. Gemini 2.5 Flash is used as the Judge LLM for statement extraction and evidence verification.

% \paragraph{Note.}
% We retain the same evaluation formulas and dimension definitions as in the original work,
% while applying them to our modified dataset.

% -----------------
\section{Failure Taxonomy Construction Pipeline}
\label{appendix:taxonomy-pipeline}

This appendix formalizes the human--machine collaborative pipeline for constructing
the failure taxonomy. It includes input specifications, a workflow overview, and
algorithmic pseudocode for reproducibility.

\subsection{Parameters}
\label{app:inputs}

\begin{center}
\begin{tabular}{@{}l p{0.8\linewidth}@{}}
\toprule
\textbf{Symbol} & \textbf{Description} \\
\midrule
$\mathbf{D}$ & Execution records collected from nine evaluated models (see \autoref{tab:overall-results}), excluding OpenManus and WebThinker. \\
$\mathbf{M}$ & A set of five LLM coders $\{ m_1, \dots, m_5 \}$, each representing a distinct model family, including Claude Opus-4.1, Gemini-2.5-Pro, Grok-4, DeepSeek-V3.1, and Qwen3-Max-Preview. \\
$\mathbf{S}_0$ & Seed concepts extracted from prior literature, used to construct few-shot prompts. \\
$\theta_{\text{sim}}$ & Cosine similarity threshold, set to $0.6$. \\
$\tau_{\text{freq}}$ & Frequency pruning threshold applied during concept filtering. \\
\bottomrule
\end{tabular}
\end{center}

\raggedbottom 
\subsection{Overview of the Pipeline}
\label{app:pipeline-overview}
\[
\textbf{Pipeline}(\mathbf{D},\mathbf{M},\mathbf{S}_0,\mathbf{P},\theta_{\text{sim}},\tau_{\text{freq}})
\rightarrow (\mathbf{C}^{\star},\mathbf{A}^{\star},\mathbf{K}^{\star})
\]
\vspace*{-1.2em}

\begin{enumerate}[label=\arabic*., leftmargin=2em, itemsep=0pt, topsep=0pt, parsep=0pt]
    \item Partition $\mathbf{D}$ into two subsets, $\mathbf{D}_A$ and $\mathbf{D}_B$ (see \autoref{tab:group_comparison}).
    \item For each subset, run \texttt{OpenCodingGen}, followed by two iterations of \texttt{OpenCodingOpt}.
    \item Merge and refine the two codebooks once to obtain $\mathbf{C}^{\star}$ (51 conceptual categoties).
    \item Perform three rounds of \texttt{AxialCodingWithICR} to derive $\mathbf{A}^{\star}$ (14 axial categories).
    \item Apply \texttt{SelectiveCoding} to abstract $\mathbf{A}^{\star}$ into the three core dimensions
          $\mathbf{K}^{\star}$ (3 core categories).
\end{enumerate}

\raggedbottom 
\subsection{Algorithmic Procedures}
\label{app:algorithms}

\subsubsection{Algorithm 1: Open Coding - Generation Stage}
\label{alg:open}
\begin{algorithm}[H]
\caption{Open Coding - Generation Stage}
\begin{algorithmic}[1]
\Procedure{OpenCodingGen}{$D_{\text{group}}, M, S_0$}
\State Initialize codebook $C \gets S_0$
\For{each execution record $e \in D_{\text{group}}$}
    \State $r \gets \text{LLM\_generate\_failure\_report}(e)$ \Comment{supplementary report}
    \For{each coder $m \in M$}
        \State $A_m \gets \text{LLM\_open\_code}(e,r,C)$
        \For{each annotation $a \in A_m$}
            \State $(name, desc) \gets \text{Normalize}(a)$
            \If{$name \in C$}
                \State $C[name].freq \gets C[name].freq + 1$
                \State $C[name].sources \gets C[name].sources \cup \{id(e), id(m)\}$
            \Else
                \State $C[name] \gets \{desc, freq=1, sources=\{id(e), id(m)\}\}$
            \EndIf
        \EndFor
    \EndFor
\EndFor
\State \Return $C$
\EndProcedure
\end{algorithmic}
\end{algorithm}

\begin{table}[h!]
\centering
\renewcommand{\arraystretch}{1.2}
\setlength{\tabcolsep}{9pt}
\resizebox{\textwidth}{!}{%
\begin{tabular}{@{} c c c c c c c @{}} 
\toprule
\textbf{Group} & \textbf{DRAs} & \textbf{Coding Model} & \textbf{Generation} & \textbf{Refinement-1} & \textbf{Refinement-2} & \textbf{Refinement-3} \\
\midrule

\multirow{5}{*}{Group A}
& \multirow{5}{*}{
\begin{tabular}[c]{@{}c@{}}
OWL\cite{hu2025owloptimizedworkforcelearning} \\
Perplexity Deep Research\cite{perplexity-deep-research}\\
MiroFlow\cite{2025mirothinker}
\end{tabular}
}
& DeepSeek-V3.1\cite{deepseek_v31_nvidia_nim_2025}     & 197 & 21 & \multirow{5}{*}{39} & \multirow{10}{*}{51} \\
& & Grok-4\cite{grok4_api_2025}            & 17  & 8  &                     &                      \\
& & Claude Opus-4.1\cite{claude_opus_41_2025}        & 17  & 11 &                     &                      \\
& & Qwen3-Max-Preview\cite{qwen3_max_preview_api_2025} & 19  & 12 &                     &                      \\
& & Gemini-2.5-Pro\cite{gemini25pro_modelcard_2025}        & 125 & 21 &                     &                      \\
\cmidrule(lr){1-6}

\multirow{5}{*}{Group B}
& \multirow{5}{*}{
\begin{tabular}[c]{@{}c@{}}
MiroThinker\cite{miromind2025mirothinker} \\
Gemini-2.5-Pro Deep Research\cite{google-gemini-deep-research} \\
O3 Deep Research\cite{openai_o3_deep_research} \\
O4-Mini Deep Research\cite{openai_o4_mini_deep_research} \\
AFM\cite{li2025chainofagentsendtoendagentfoundation}
\end{tabular}
}
&  DeepSeek-V3.1\cite{deepseek_v31_nvidia_nim_2025}     & 477 & 12 & \multirow{5}{*}{29} &                      \\
& & Grok-4\cite{grok4_api_2025}           & 29  & 16 &                     &                      \\
& & Claude Opus-4.1\cite{claude_opus_41_2025}        & 109 & 17 &                     &                      \\
& & Qwen3-Max-Preview\cite{qwen3_max_preview_api_2025} & 364 & 14 &                     &                      \\
& & Gemini-2.5-Pro\cite{gemini25pro_modelcard_2025}       & 214 & 17 &                     &                      \\
\bottomrule
\end{tabular}
}% end resizebox
\caption{Comparison of model generations and refinements between Group A and Group B.}
\label{tab:group_comparison}
\end{table}

% ============================================================
\subsubsection{Algorithm 2: Open Coding - Optimization Stage}
\label{alg:opt}

% We first form candidate pairs by cosine similarity:
% \[
% \text{if } \cos(\mathrm{Emb}(u),\mathrm{Emb}(v)) \ge \theta_{\text{sim}},
% \ \text{then } (u,v)\in \mathcal{P}.
% \]

% \begin{algorithm}[H]
% \caption{Open Coding - Optimization Stage}
% \begin{algorithmic}[1]
% \Procedure{OpenCodingOpt}{$C, \theta_{\text{sim}}, \tau_{\text{freq}}$}
% \For{$r = 1$ to \text{rounds}}
%     \State $\text{Pairs} \gets \text{FindSimilarPairs}(C, \theta_{\text{sim}})$
%     \State $\text{Groups} \gets \text{UnionFind}(\text{Pairs})$
%     \For{each group $G$ in Groups}
%         \State $C_{\text{merged}} \gets \text{LLM\_merge\_concepts}(\{C[g] \mid g \in G\})$
%         \State Remove all $g \in G$ from $C$; Insert $C_{\text{merged}}$ into $C$
%     \EndFor
%     \For{each $c \in C$}
%         \If{$c.\text{freq} < \tau_{\text{freq}}$}
%             \State Remove $c$
%         \EndIf
%     \EndFor
% \EndFor
% \State \Return $C$
% \EndProcedure
% \end{algorithmic}
% \end{algorithm}

\begin{algorithm}[H]
\caption{Open Coding -- Optimization Stage}
\begin{algorithmic}[1]
\Procedure{OpenCodingOpt}{$C, \theta_{\text{sim}}, \tau_{\text{freq}}$}

\State $changed \gets \text{true}$
\While{$changed$}
    \State $changed \gets \text{false}$
    \State $best\_pair \gets \text{null}$
    \State $max\_sim \gets -1$
    \For{each $c_i$ in $C$}
        \For{each $c_j$ in $C$ with $j > i$}
            \State $sim \gets \text{CosineSimilarity}(c_i, c_j)$
            \If{$sim > \theta_{\text{sim}}$ \textbf{and} $sim > max\_sim$}
                \State $max\_sim \gets sim$
                \State $best\_pair \gets (c_i, c_j)$
            \EndIf
        \EndFor
    \EndFor
    \If{$best\_pair \neq \text{null}$}
        \State $(c_1, c_2) \gets best\_pair$
        \State $merged \gets \text{LLM\_merge\_concepts}(c_1.\text{name}, c_1.\text{desc}, c_2.\text{name}, c_2.\text{desc})$
        \If{$merged \neq \text{null}$}
            \State $C \gets C \setminus \{c_1, c_2\}$
            \State $C \gets C \cup \{merged\}$
            \State $changed \gets \text{true}$
        \EndIf
    \EndIf
\EndWhile

\For{each $c \in C$}
    \If{$c.\text{freq} < \tau_{\text{freq}}$}
        \State $C \gets C \setminus \{c\}$
    \EndIf
\EndFor

\State \Return $C$
\EndProcedure
\end{algorithmic}
\end{algorithm}

% ============================================================
% \subsubsection{Algorithm 3: Cross-group Integration}
% \label{alg:merge}

% \begin{algorithm}[H]
% \caption{Cross-group Integration}
% \begin{algorithmic}[1]
% \Procedure{MergeAcrossGroups}{$C_A, C_B, \theta_{\text{sim}}, \tau_{\text{freq}}$}
% \State $C_{\text{union}} \gets C_A \cup C_B$
% \State $C_{\star} \gets \text{OptimizeConcepts}(C_{\text{union}}, \theta_{\text{sim}}, \tau_{\text{freq}}, \text{rounds}=1)$
% \State \Return $C_{\star}$ \Comment{ 51 concepts}
% \EndProcedure
% \end{algorithmic}
% \end{algorithm}

% ============================================================
\subsubsection{Algorithm 3: Axial Coding with ICR Evaluation}
\label{alg:axial}

Each iteration examines semantic, contextual, processual, causal, functional,
structural, and strategic relationships among concepts. Inter-coder reliability (ICR)
is assessed using Krippendorff’s $\alpha = 1 - D_o / D_e$ on stratified samples of
24 (Round~1) and 54 (Rounds~2--3) records annotated independently by three domain experts,
followed by reconciliation sessions of approximately five hours each.

% \begin{algorithm}[H]
% \caption{Axial Coding with ICR Evaluation}
% \begin{algorithmic}[1]
% \Procedure{AxialCodingWithICR}{$C_{\star}, D$}
% \For{$t \in \{1, 2, 3\}$}
%     \State $\text{Base} \gets$ 
%         \textbf{if} $t == 1$ \textbf{then} ConceptsFromGroupA($C_{\star}$) 
%         \textbf{else} ($C_{\star} \cup A_{\text{prev}}$)
%     \State $A_t \gets \text{LLM\_axial\_coding}(\text{Base}, 
%         \text{criteria} = [\text{semantic, context, process, causal, functional, structural, strategic}])$
%     \State $n \gets 24$ \textbf{if} $t==1$ \textbf{else} $54$
%     \State $S_t \gets \text{StratifiedSample}(D, n)$
%     \State $\text{Labels} \gets \{ \text{expert}_j : \text{ExpertLabel}(S_t, A_t) \text{ for } j=1..3 \}$
%     \State $\alpha \gets \text{KrippendorffAlpha}(\text{Labels})$
%     \State $A_t \gets \text{ExpertDiscussionRefine}(A_t, \text{Labels}, \alpha)$
%     \State $A_{\text{prev}} \gets A_t$
% \EndFor
% \State $A_{\star} \gets \text{EnforceTargetCount}(A_{\text{prev}}, \text{target}=14)$
% \State \Return $A_{\star}$
% \EndProcedure
% \end{algorithmic}
% \end{algorithm}
\begin{algorithm}[H]
\caption{Axial Coding with ICR Evaluation}
\begin{algorithmic}[1]
\Procedure{AxialCodingWithICR}{$C^{\star}, D$}
\For{$t \in \{1, 2, 3\}$}
    \If{$t == 1$}
        \State $\text{Base} \gets \text{ConceptsFromGroupA}(C^{\star})$
    \Else
        \State $\text{Base} \gets (C^{\star} \cup A_{\text{prev}})$
    \EndIf
    \State $A_t \gets \text{Human\_LLM\_axial\_coding}(\text{Base, criteria=}$
    \Statex \hspace{2em} $[\text{semantic, context, process, causal, functional, structural, strategic}])$
    \State $n \gets 24$ \textbf{ if } $t == 1$ \textbf{ else } $54$
    \State $S_t \gets \text{StratifiedSample}(D, n)$
    \State $\text{Labels} \gets \{\text{expert}_j : 
        \text{ExpertLabel}(S_t, A_t) \text{ for } j = 1..3 \}$
    \State $\alpha \gets \text{KrippendorffAlpha}(\text{Labels})$
    \State $A_t \gets \text{ExpertDiscussionRefine}(A_t, \text{Labels}, \alpha)$
    \State $A_{\text{prev}} \gets A_t$
\EndFor
\State \Return $A^{\star}$
\EndProcedure
\end{algorithmic}
\end{algorithm}

% ============================================================
\subsubsection{Algorithm 4: Selective Coding}
\label{alg:selective}
\begin{algorithm}[H]
\caption{Selective Coding}
\begin{algorithmic}[1]
\Procedure{SelectiveCoding}{$A^{\star}$}
\State $K \gets \text{Human\_LLM\_selective\_coding}(A^{\star})$
\State $\text{Relations} \gets \text{BuildClosedLoop}(K)$ \Comment{temporal progression + functional cycle}
\State \Return $\{ K, \text{Relations} \}$
\EndProcedure
\end{algorithmic}
\end{algorithm}

\noindent
The final output $\mathbf{K}^{\star}$ provides a three-dimensional view that captures
cognitive, retrieval, and generative aspects of failure. This hierarchical structure
supports transparent error analysis and reproducible categorization across datasets and models.

\section{Seed Conceptual Categories} \label{app:seed}

This appendix provides three seed conceptual categories used to guide open coding of LLM.
\begin{center}
\begin{tcolorbox}[
  colback=yellow!10!white,    % 背景亮黄
  colframe=orange!70!black,   % 边框橙黄混合
  colbacktitle=orange!60!black, % 标题栏橙黄
  coltitle=white,             % 白色标题文字
  fonttitle=\bfseries,
  title=Seed Conceptual Categories,
  width=1\linewidth,
  boxrule=0.8pt,
  enhanced,
  %%%sharp corners,
  breakable
]

\begin{description}
\item [\textbf{Failure to Understand Requirements}:] The agent fails to correctly interpret user requirements, intent, or contextual needs, focusing on superficial keyword matches rather than the actual problem, resulting in responses that don't align with the user's goals.
\item [\textbf{Information Retrieval Bypass}:] The agent generates content from internal knowledge rather than performing actual external information retrieval when the task explicitly requires collecting existing materials.
\item [\textbf{Format and Structural Non-Compliance}:] The agent failed to follow the user specified output structure, layout format, or presentation specifications, affecting professional delivery and information readability.

\end{description}
\end{tcolorbox}
\end{center}

\section{Computation of Krippendorff’s Alpha}
\label{app:alpha}

\subsection{Data and Scope}
Krippendorff’s $\alpha$ was computed to assess inter-coder reliability between human experts and the LLM (\textit{Gemini 2.5 Flash}) across three categories of coded items. 
A total of 14 items were included: 4 in the \textit{Reasoning} category, 5 in \textit{Retrieval}, and 5 in \textit{Generation}.

Two levels of coefficients were derived:
\begin{itemize}
    \item \textbf{Overall $\alpha$ (overall\_alpha):} Computed across all 14 items, reflecting the overall consistency of coding, including cross-category variation.
    \item \textbf{Category-level $\alpha$ (category\_alphas):} Computed within each category subset, reflecting intra-category consistency only.
\end{itemize}

Because the expected disagreement term $D_e$ depends on the marginal distribution of categories, the overall $\alpha$ is \textit{not} a simple or weighted average of the category-level coefficients.

\subsection{Formal Definition}
For nominal data, Krippendorff’s $\alpha$ is defined as:
\[
\alpha = 1 - \frac{D_o}{D_e}, \qquad
D_o = \frac{\sum_{c}\sum_{k \neq c} n_{ck} \, \delta(c,k)}{\sum_{c} n_c (n_c - 1)}, \quad
D_e = \frac{\sum_{c}\sum_{k \neq c} N_c N_k \, \delta(c,k)}{N (N - 1)},
\]
where $\delta(c,k)=1$ if $c \neq k$ and $0$ otherwise.  
The \textbf{overall\_alpha} aggregates all 14 items when computing $D_o$ and $D_e$, while the \textbf{category\_alphas} are computed within each subset. Hence, when inter-category variance is large, the overall $\alpha$ may diverge from the category-level estimates.

\subsection{Computation Settings}
% \begin{itemize}
%     \item \textbf{Measurement level:} Nominal (discrete categorical labels).  
%     \item \textbf{Missing values:} Allowed; no post-hoc adjudication performed.  
%     \item \textbf{Estimation:} Conducted using the Python \texttt{krippendorff} package with 1{,}000 bootstrap iterations to obtain 95\% confidence intervals.
% \end{itemize}

\begin{table}[h!]
\centering
\caption{Summary of computation settings for Krippendorff's $\alpha$}
\label{tab:alpha_settings}
\renewcommand{\arraystretch}{1.3}
\begin{tabular}{@{}ll@{}}
\toprule
\textbf{Aspect} & \textbf{Description} \\
\midrule
Measurement level & Nominal (discrete categorical labels) \\
Missing values & Allowed; no post-hoc adjudication performed \\
Estimation method & Python \texttt{krippendorff} package with 1{,}000 bootstrap iterations \\
Output & Point estimates of $\alpha$ for each category and the overall dataset \\
\bottomrule
\end{tabular}
\end{table}

% /
The overall coefficient reflects agreement across all items, incorporating both intra- and inter-category variation. In contrast, the category-level coefficients isolate agreement within each conceptual dimension. The difference between these estimates provides insight into how cross-category variance affects overall coding reliability. A high $\alpha$ (above 0.80) across both levels indicates strong coder consistency and conceptual clarity of the \bench{} framework.

\section{FINDER Stability Analysis via MiroFlow}
\label{app:miroflow}
To assess the stability and cross-lingual consistency of \textbf{\bench{}}, we perform a multi-run evaluation using \textbf{MiroFlow} as a representative agent framework. MiroFlow is selected because it attains the highest overall performance among the evaluated frameworks on \bench{}, making it a suitable testbed for stability analysis. We conduct three independent runs with both English (EN) and Chinese (ZH) prompts. The raw and aggregated \textbf{RACE} results are summarized in \autoref{tab:miroflow_race_summary}.

As shown in \autoref{tab:miroflow_race_summary}, the standard deviations across runs are small, indicating stable \textbf{\bench{}} performance under repeated trials and across languages. The EN setting achieves slightly higher mean scores on \textit{Overall}, \textit{Comprehensiveness}, and \textit{Depth}, suggesting modestly stronger reasoning and content generation in English. In contrast, \textit{Instruction-following} and \textit{Readability} are nearly identical between EN and ZH prompts, demonstrating consistent instruction adherence and output fluency across languages.

\begin{table}[h!]
\centering
\caption{MiroFlow RACE Results Summary (Three Runs)}
\label{tab:miroflow_race_summary}
\normalsize
\renewcommand{\arraystretch}{1.15}
\setlength{\tabcolsep}{7pt}
\begin{tabular}{lcccc}
\toprule
\textbf{Dimension} & \textbf{EN Mean} & \textbf{EN Std} & \textbf{ZH Mean} & \textbf{ZH Std} \\
\midrule
Overall & 45.54 & 0.43 & 44.49 & 0.20 \\
Comp.   & 45.58 & 0.63 & 44.43 & 0.16 \\
Depth   & 41.63 & 0.56 & 39.16 & 0.36 \\
Inst.   & 49.61 & 0.26 & 49.35 & 0.17 \\
Read.   & 46.61 & 0.09 & 46.86 & 0.09 \\
\bottomrule
\end{tabular}
\end{table}

\begin{figure}[h!]
\centering
\includegraphics[width=0.7\linewidth]{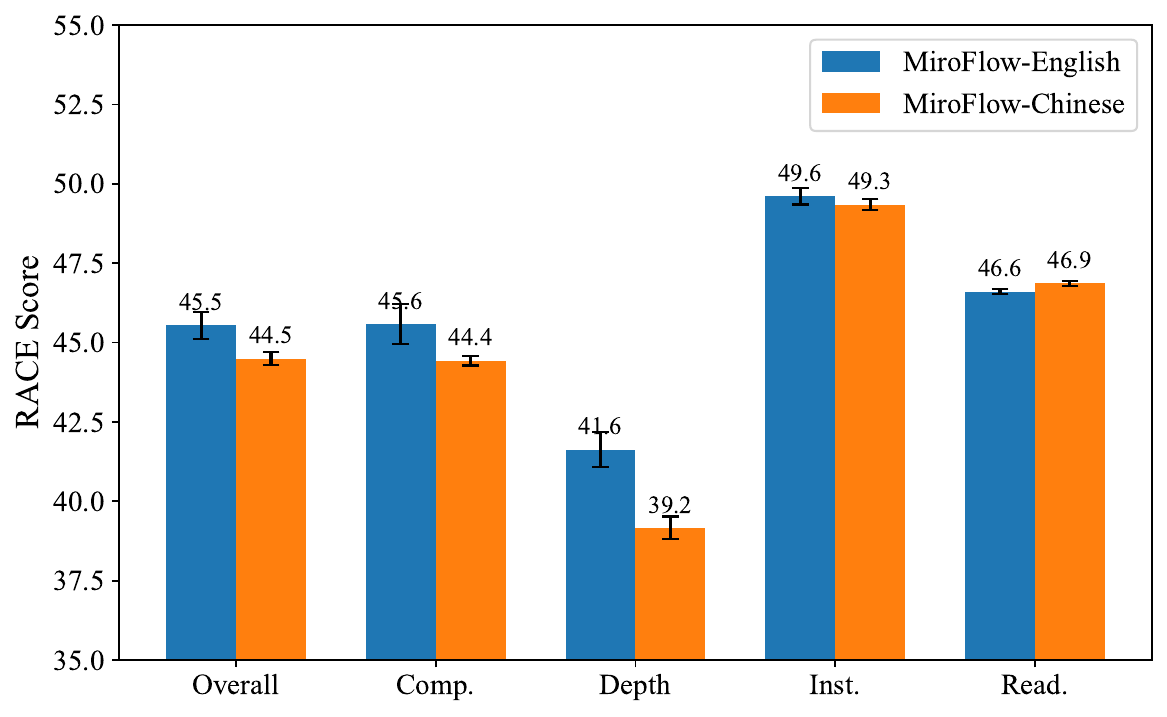}
\caption{Comparison of FINDER RACE Results under MiroFlow (EN vs. ZH; mean over three runs)}
\label{fig:miroflow_race_bar}
\end{figure}

\autoref{fig:miroflow_race_bar} visualizes the mean RACE scores with corresponding standard deviations. EN prompts yield consistently but only marginally higher scores, whereas both EN and ZH settings exhibit strong run-to-run stability, further supporting the robustness of \textbf{\bench{}} in multilingual scenarios.

% ----------------
\section{Checklist Distribution}
\label{app:checklist_distribution}

% \textbf{Figure \ref{fig:checklist_distribution}} shows that among all the queries, the number of checklists varies from 3 to 5, with 5 checklists being the most numerous, accounting for half of the total queries.

\begin{figure}[H]
    \centering
    \includegraphics[width=0.6\textwidth]{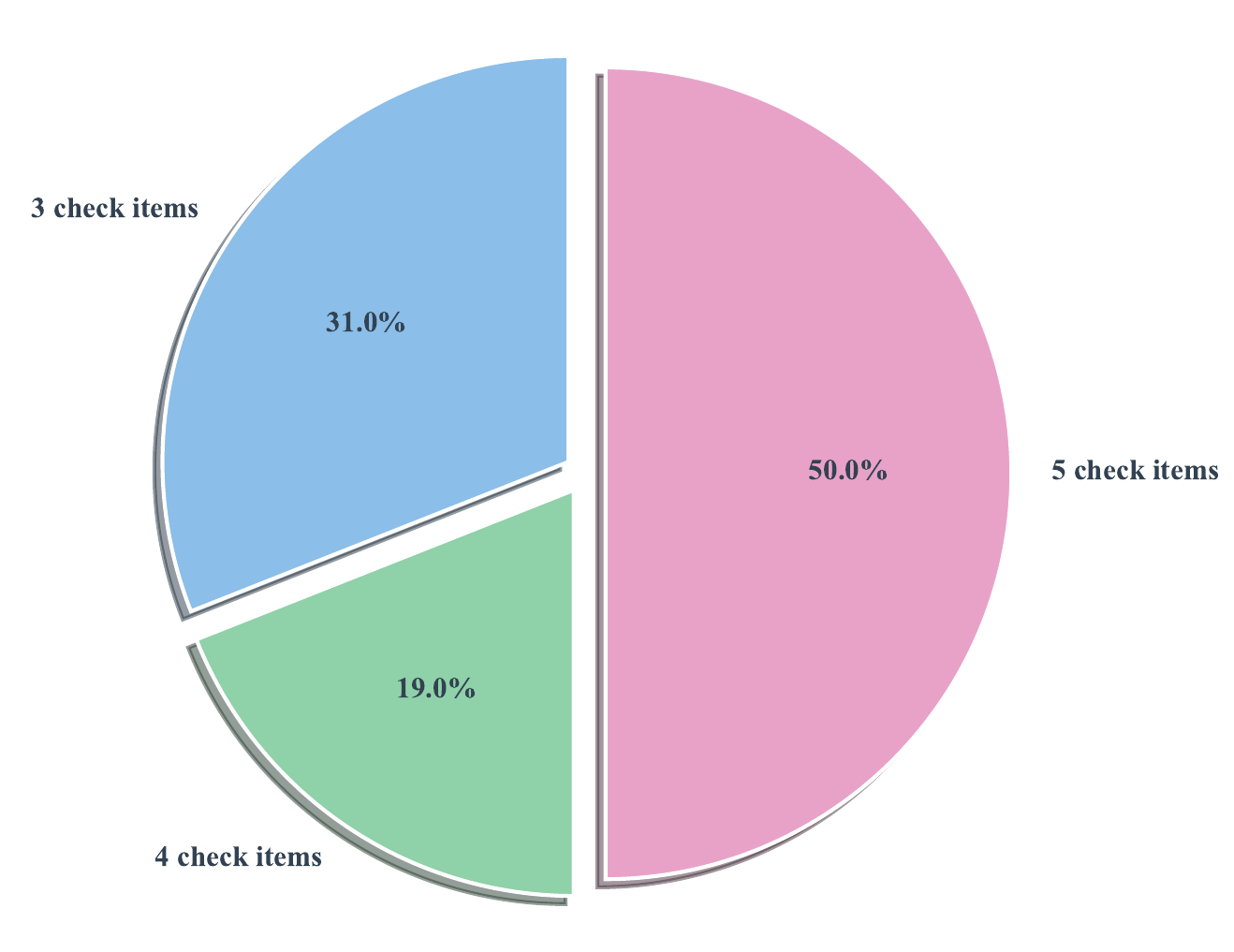}
    \caption{The distribution of checklists in queries of \bench{}}
    \label{fig:checklist_distribution}
\end{figure}

\section{Configuration of Evaluated Models}
\label{app:model_config}

This appendix summarizes the configurations of all evaluated models used in our experiments. 
All models were used with their default system prompts and inference parameters unless otherwise stated.

\subsection{Proprietary API Models}

These models were accessed through their official APIs with default configurations. 
No tool integration or parameter tuning was applied.

\subsection{Open-source Models}

\begin{itemize}
    \item \textbf{WebThinker}: \texttt{WebThinker-QwQ-32B}, with integrated web search. Default parameters used.
    \item \textbf{AFM}: \texttt{AFM-45B}, a multi-agent academic reasoning system with citation verification. Key parameters: \texttt{temperature = 0.4}, \texttt{top\_p = 0.9}, \texttt{max\_tokens = 32K}.
    \item \textbf{MiroThinker}: \texttt{MiroThinker-32B-DPO-v0.2} with \texttt{max\_tokens = 64K}, using a multimodal vision model (\texttt{Qwen2.5-VL-\allowbreak72B-Instruct}) for image inputs.
\end{itemize}

\subsection{Agent Frameworks}

\begin{itemize}
    \item \textbf{MiroFlow}: Dual-agent framework based on \texttt{Claude-3.7-Sonnet}. Key parameters: \texttt{temperature = 0.3}, \texttt{top\_p = 0.95}, \texttt{max\_tokens = 32K}.
    \item \textbf{OWL}: Multi-agent architecture powered by \texttt{OpenAI O1}, integrating modules for reasoning, planning, and multimodal perception.
    \item \textbf{OpenManus}: \texttt{gpt-4o}-based agent system with automated web and code execution tools. Key parameters: \texttt{temperature = 0.0}, \texttt{max\_tokens = 8192}.
\end{itemize}

\section{Positive Taxonomy Metric}
\label{app:cosine-metric}

This appendix provides a detailed justification for the positive-taxonomy scoring metric used in our analysis. Let $|D|$ denote the total number of evaluated instances and let $E_i \in [0, |D|]$ be the error count associated with category~$i$. The metric is defined as
\begin{equation}
S_i = |D| \cdot \cos\!\left(\frac{E_i}{|D|}\cdot\frac{\pi}{2}\right).
\end{equation}
a cosine-based transformation mapping error counts into a bounded, positive scale.

The function is strictly monotonic decreasing and invertible over the domain $E_i\in[0,|D|]$, ensuring that it preserves all information contained in the raw error counts while providing a normalized and interpretable representation. This behavior makes it a suitable reparameterization for analyzing model performance across taxonomy categories.

A key motivation for adopting the cosine form lies in its curvature. Near the low-error regime ($E_i\approx 0$), the curve is relatively flat, meaning that very small increases in error induce minimal reductions in score. This reflects our analytical preference not to over-emphasize distinctions among categories that already exhibit highly reliable performance. As error increases, however, the curve becomes progressively steeper, producing sharper declines in score. This naturally concentrates resolution in the portions of the error range where error rates reach moderate and higher levels, making differences between categories more diagnostically significant. A linear mapping, by contrast, has constant slope and cannot provide this targeted emphasis.

The metric further benefits from an intuitive interpretability analogy. Inspired by classical cosine similarity in information retrieval~\citep{10.1145/361219.361220}, the score $S_i$ may be viewed as measuring the angular deviation between performance in category~$i$ and an ideal, error-free direction. A category with zero errors aligns perfectly with this ideal, yielding a maximal score; larger error rates correspond to larger angular deviations and therefore smaller cosine values. This interpretation provides a geometric perspective on category-level performance that aligns closely with intuitive notions of similarity to a reference model.

\section{Analysis of Missing Results of FACT framework}
\label{appendix:fact_failure}

During evaluation, we found that several models failed to produce valid outputs within the FACT framework. A follow-up analysis indicates that these failures fall into four primary categories. This appendix systematically examines each category to clarify potential sources of evaluation bias and to delineate limitations inherent to the FACT framework.

\begin{itemize}

    \item \textbf{Anti-Scraping Mechanisms.}  
    Many academic publishers, government agencies, and commercial websites employ anti-scraping protections. Consequently, Jina AI Reader often cannot access or parse these pages. This results in missing citations and incomplete retrieval chains, thereby weakening the reliability of FACT scores.

    \begin{quote}
    \colorbox{gray!10}{\parbox{\linewidth}{
    \noindent\textbf{Examples:}
    \begin{itemize}
        \item \texttt{https://www.tandfonline.com/doi/full/10.1080/14780887.2020.1769238\#abstract}
        \item \texttt{https://onlinelibrary.wiley.com/doi/10.1207/s15516709cog1202\_4}
        \item \texttt{https://www.tandfonline.com/doi/abs/10.1207/S15327965PLI1104\_01}
        \item \texttt{https://ingenico.com/us-en}
        % \item \texttt{https://www.sciencedirect.com/science/article/pii/S0735109720358009}
        % \item \texttt{https://www.nejm.org/doi/full/10.1056/NEJMoa0908355}
        % \item \texttt{https://www.iata.org/}
        % \item \texttt{https://www.imf.org/en/Home/}
    \end{itemize}
    }}
    \end{quote}

    \item \textbf{Non-Existent or Fabricated URLs.}  
    In some instances, models generated URLs that do not correspond to real webpages. Such failures are typically caused by hallucinated links or by broader model limitations, which prevent the retrieval system from accessing the intended content.

    \begin{quote}
    \colorbox{gray!10}{\parbox{\linewidth}{
    \noindent\textbf{Examples:}
    \begin{itemize}
        \item \texttt{http://moe.gov.cn/}
        \item \texttt{http://gd.gov.cn/}
        \item \texttt{https://go.isi/mda2}
    \end{itemize}
    }}
    \end{quote}

    \item \textbf{Incorrect URL Formats.}  
    Some model outputs include academic references or citation strings that resemble URLs. Because of the internal URL-extraction rules in \texttt{deep\_research\_bench}, these strings may be misidentified as valid links. Since they do not map to actual web resources, retrieval fails by design.

    \begin{quote}
    \begin{tcolorbox}[colback=gray!10, colframe=gray!10, boxrule=0pt, arc=2pt, left=4pt, right=4pt, top=4pt, bottom=4pt]
    \noindent\textbf{Example error log:}
    \begin{Verbatim}[fontsize=\small]
ERROR: Failed to fetch Porter, M. & Heppelmann, J. (2021).
"Digital twins in critical water infrastructure".
IEEE Engineering Management Review, 49(3), 72–81.
Jina AI Reader Failed ... 400
    \end{Verbatim}
    \end{tcolorbox}
    \end{quote}

    \item \textbf{Timeout and Rate Limiting Issues.}  
    Retrieval failures can also arise from system-level constraints, including network latency, high request volume, or temporary API throttling. These conditions may trigger timeouts, preventing content from being returned within the evaluation. As shown in the table below, successive versions exhibited only minor changes, and Krippendorff’s Alpha steadily increased across iterations.

\end{itemize}

\end{document}